\def\bm{{\mathbf m}}

\def\bI{{\mathbf I}}

\def\texitem#1{\par\smallskip\noindent\hangindent 25pt
               \hbox to 25pt {\hss #1 ~}\ignorespaces}

\def\c{{\mathsf{c}}}

\documentclass[fleqn,10pt]{wlscirep}

\usepackage{amsmath,amssymb}
\usepackage{amsmath,amsfonts,amsthm,bm}

\usepackage{changepage}

\usepackage{textcomp,marvosym}

\usepackage{nameref,hyperref}

\usepackage[right]{lineno}

\usepackage{subcaption}

\usepackage{microtype}
\DisableLigatures[f]{encoding = *, family = * }

\usepackage{array}

\usepackage{booktabs}

\newcolumntype{+}{!{\vrule width 2pt}}

\newlength\savedwidth

\usepackage{lineno,hyperref}

\usepackage{comment}
\usepackage[utf8]{inputenc}
\usepackage[T1]{fontenc}
\title{Combining imitation and deep reinforcement learning to human-level performance on a virtual foraging task}

\author[1,*]{Vittorio Giammarino}
\author[4,5,6]{Matthew F Dunne}
\author[4,5,6]{Kylie N Moore}
\author[6]{Michael E Hasselmo}
\author[4,5]{Chantal E Stern}
\author[1,2,3]{Ioannis Ch. Paschalidis}

\affil[1]{Division of Systems Engineering, Boston University, Boston, MA 02446, USA.}
\affil[2]{Dept. of Electrical and Computer Engineering, Boston University, Boston, MA 02446, USA.}
\affil[3]{Dept. of Biomedical Engineering, Boston University, Boston, MA 02215, USA.}
\affil[4]{Cognitive Neuroimaging Center, Boston University, Boston, MA 02215, USA.}
\affil[5]{Graduate Program for Neuroscience, Boston University, Boston, MA 02215, USA.}
\affil[6]{Center for Systems Neuroscience, Boston University, Boston, MA 02215, USA.}

\affil[*]{vgiammar@bu.edu}

\begin{abstract}
We develop a simple framework to learn bio-inspired foraging policies using human data. We conduct an experiment where humans are virtually immersed in an open field foraging environment and are trained to collect the highest amount of rewards. A Markov Decision Process (MDP) framework is introduced to model the human decision dynamics. Then, Imitation Learning (IL) based on maximum likelihood estimation is used to train Neural Networks (NN) that map human decisions to observed states. The results show that passive imitation substantially underperforms humans. We further refine the human-inspired policies via Reinforcement Learning (RL) using the on-policy Proximal Policy Optimization (PPO) algorithm which shows better stability than other algorithms and can steadily improve the policies pretrained with IL. We show that the combination of IL and RL can match human results and that good performance strongly depends on combining the allocentric information with an egocentric representation of the environment. 
\end{abstract}
\begin{document}

\flushbottom
\maketitle
\thispagestyle{empty}

\section*{Introduction}

Human beings are exceptional learners: capable of conceiving solutions for individual problems, generalizing acquired skills to new tasks, exploring new strategies, and inferring causal relationships. \cite{walker2012explaining, gopnik2015younger, goddu2020transformations, ruggeri2021toddlers} Since the earliest stage of Machine Learning (ML), the research community has sought to emulate humans' learning capacities; and only recently, works which combine Deep Learning (DL) with Reinforcement Learning (RL), have accomplished outstanding results in this regard.\cite{silver2017mastering} DL and RL have shown to be indispensable ingredients to accomplish human-like intelligence in artificial systems; however, they still require a massive amount of computational resources and do not show the same level of efficiency compared to human beings.\cite{botvinick2019reinforcement} A viable option to tackle the efficiency issue, again inspired by human learning,\cite{offerman1998learning,jones2009development} is to leverage human demonstrations by combining DL and RL with imitation in a procedure known as imitation learning (IL).\cite{pomerleaualvinn} It is worth noting that a great number of tasks, such as navigating or exploring unknown environments, are relatively straightforward for humans and can be successfully learned in a limited number of trials. On the other hand, this is often not the case for artificial agents, where the amount and the quality of the information retrieved, in addition to a sound design of the reward function and/or a good exploration strategy of the environment, are crucial for successfully learning from scratch. Hence, artificial agents might benefit from directly imitating human behavior or, alternatively, from reconstructing human-inspired reward functions.\cite{abbeel2010autonomous, uchibe2021forward}  

In this study, we investigate the potential of learning from humans taking into account not only performance but also efficiency. We start by collecting movement data from a series of human participants while they are performing a virtual foraging task in which the rewards, in the form of coins, are condensed in clusters throughout the environment. The participants, subject to time constraints, have to collect the highest number of coins, effectively trading-off between foraging within a single cluster (exploitation) and exploration. Humans are initially unaware of the number of clusters and of their locations but are able to learn the reward distribution throughout the course of the experiment.

Note that, time-constrained foraging problems occur in several realistic scenarios, including scientific exploration; where, for instance, a rover might want to sample chemical or geological features as fast as possible; or search and rescue operations, where a vessel needs to rescue as many people as possible.\cite{scone2010trade,otte2013navigation} Moreover, these missions are dangerous, and the use of aerial or ground unmanned vehicles would significantly mitigate any additional risks. However, without assumptions about the distribution of targets, classical control techniques are not applicable. Tele-operation is also a feasible option, but it may be hindered due to unreliable communications, and this approach does not scale as well as completely autonomous options. For these reasons, ML techniques supporting full autonomy represent an interesting alternative solution.\cite{ccatal2021robot} Therefore, our main objective is to develop a method which effectively combines IL with DL and RL and allows for efficient human-level learning in a foraging task with sparse rewards. 

From our experiment, we collect $50$ human trajectories and further process them to include both allocentric and egocentric information in our model. We then run IL on each of the trajectories, yielding $50$ policies with different performance. None of these policies succeed in matching human results. Finally, we use the imitated policies as initial solutions and further refine them with RL. By combining the two methods, we outperform the average human performance and the respective participant from which the agent imitates with a success rate of 78\% and 62\%, respectively, while using a reasonable amount of training steps ($\leq 10^7$). We also compare our method with a pure RL alternative, and show that such an approach remarkably underperforms humans. 

In the final part of the work, we test the learned policies for generalization and robustness in a new scenario with an unknown reward distribution. We are able to show that the artificial agents quickly adapt to this new scenario and conjecture that, when combined with allocentric information, the egocentric representation of the environment plays a key role in enabling learning as also observed in neurophysiological recordings from rodents.\cite{alexander2020egocentric} To empirically test this hypothesis, we rerun the entire set of experiments only considering allocentric coordinates. We show that in the absence of an egocentric representation of the environment, RL is unable to further improve IL and the final performance does not reach the level of human subjects. For the sake of completeness, we rerun the experiments also considering only the egocentric representation of the environment. This setup, however, violates the MDP assumption and, since our agents are not equipped with explicit memory, the learnt policies underperform those of the other experiments. A figure illustrating all these experiments for the full set of $50$ human trajectories is included in the supplementary materials. We conclude that proper modeling is as crucial as the right algorithmic choices in order to enhance general and robust learning in artificial agents.

\subsection*{Related Work and Contribution}
We focus on combining IL with RL in order to address the shortcomings of these two approaches when used individually. IL was initially proposed as a supervised learning method for faster policy learning.\cite{pomerleaualvinn, schaal1999imitation} Recent works have studied the limitation of IL including the covariate shift problem and its dependency on the quality of the demonstrations.\cite{syed2010reduction,ross2010efficient, osa2018algorithmic}. 
RL instead was proposed to enable learning through direct interaction with the environment.\cite{sutton2018reinforcement} RL combined with DL has achieved outstanding results in policy learning\cite{silver2017mastering}, however, sample inefficiency remains an obstacle for its deployment in real world scenarios.\cite{dulac2019challenges} Recent works have endeavored to combine IL with RL either to address the limitations of IL\cite{ross2011reduction, ross2014reinforcement, sun2018truncated, cheng2019fast} or to improve efficiency in RL. \cite{subramanian2016exploration, vecerik2017leveraging, nair2018overcoming, libardi2021guided, Uchendu2021DemonstrationGuidedQ} 

Another line of research, known as Inverse Reinforcement Learning (IRL), leverages demonstrations in order to infer a reward function which is then used for RL in order to recover the demonstrator policy.\cite{abbeel2004apprenticeship, ratliff2006maximum, ziebart2008maximum, finn2016guided} IRL has the pros of being immune to the covariate shift problem but the cons of being as sample inefficient as the used RL algorithm. A unified view of IRL and IL as an $f$-divergence minimization problem has been recently proposed\cite{ho2016generative,ghasemipour2020divergence} and addressed using Generative Adversarial Networks\cite{goodfellow2020generative} (GAN) for either IL\cite{ho2016generative} or reward shaping\cite{kang2018policy}.

We place our work in the Reinforcement Learning with Expert Demonstrations (RLED) framework, where the RL agent learns in the same environment of the demonstrator and using the same reward function.\cite{hester2018deep} 
Our main contribution is leveraging a non trivial case study to show how modeling, imitation and reinforcement, when effectively combined, can lead to human-like performance in navigation tasks with sparse rewards without requiring a massive amount of training steps. Note that, this does not mean that RL-only algorithms cannot achieve human-level performance in this type of tasks, rather that they need a significantly larger number of steps to do so. Dense reward signals would have most likely improved RL-only performance but also assumed prior knowledge of the environment invalidating therefore the comparison with the human agents. Moreover, we analyze our results for robustness and empirically show a strong correlation between egocentric representation of the environment and performance. We reemphasize the importance of the right algorithmic choices as well as the right model in order to enhance effective learning in artificial agents.

Two works which likewise combine IL with RL are Deep Q-Learning from Demonstrations (DQfD)\cite{hester2018deep} and AlphaGO\cite{silver2017mastering}. However, with respect to AlphaGO, which lies in the model-based spectrum, our work considers a pure model-free RL setting. DQfD, on the other hand, explores pretraining a deep Q network\cite{mnih2013playing} using demonstrations before performing RL. In order to do so, the agent needs access not only to the demonstrator state-action pairs but also to the rewards collected along the trajectories. In other words, DQfD assumes access to the full MDP transition (states, actions and rewards) and as a result, its pretrain step can be seen as a first form of offline RL \cite{levine2020offline}. In our study, we assume access only to demonstrator state-action pairs (without rewards) and therefore DQfD is not directly applicable. Furthermore, none of the aforementioned works investigate the effects of modeling on algorithmic performance.

\subsection*{Outline and Notation}
The remainder of the paper is organized as follows. The Materials and Methods section presents the experimental setup used to collect the human foraging data, introduces the MDP model used for representing human behavior, discusses the IL for learning policies from data, and outlines the RL algorithms used to refine the imitated policies. In the Results section we compare our method with human and RL-only performance on the original setup; then, we test all the artificial agents for robustness to reward distribution shift and demonstrate the importance of egocentric information. We discuss the results in the Discussion section.

\paragraph{Notation:} Unless otherwise indicated, we use uppercase letters (e.g., $S_t$) for random variables, lowercase letters (e.g., $s_t$) for values of random variables, script letters (e.g., $\mathcal{S}$) for sets, and bold lowercase letters (e.g., $\bm{\theta}$) for vectors. Let $[t_1 : t_2]$ be the set of integers $t$ such that $t_1 \leq t \leq t_2$; we write $S_t$ such that $t_1 \leq t \leq t_2$ as $S_{t_1 : t_2}$. We denote by $\mathcal{N}(\mu,\sigma^2)$ the normal distribution, where $\mu$ is the mean and $\sigma$ the standard deviation. $\bI$ denotes the identity matrix.
We denote the multivariate normal distribution with $\mathcal{N}(\bm{\mu},\sigma^2 \bI)$ where $\bm{\mu}$ is the mean vector and the covariance matrix is diagonal with only $\sigma^2$ as elements on the diagonal.  Finally, $\mathbb{E}[\cdot]$ represents expectation and $\mathbb{P}(\cdot)$ probability.

\section*{Materials and Methods}
\begin{figure}[ht]
    \centering
    \begin{subfigure}[t]{0.3\textwidth}
        \centering
        \includegraphics[width=\linewidth]{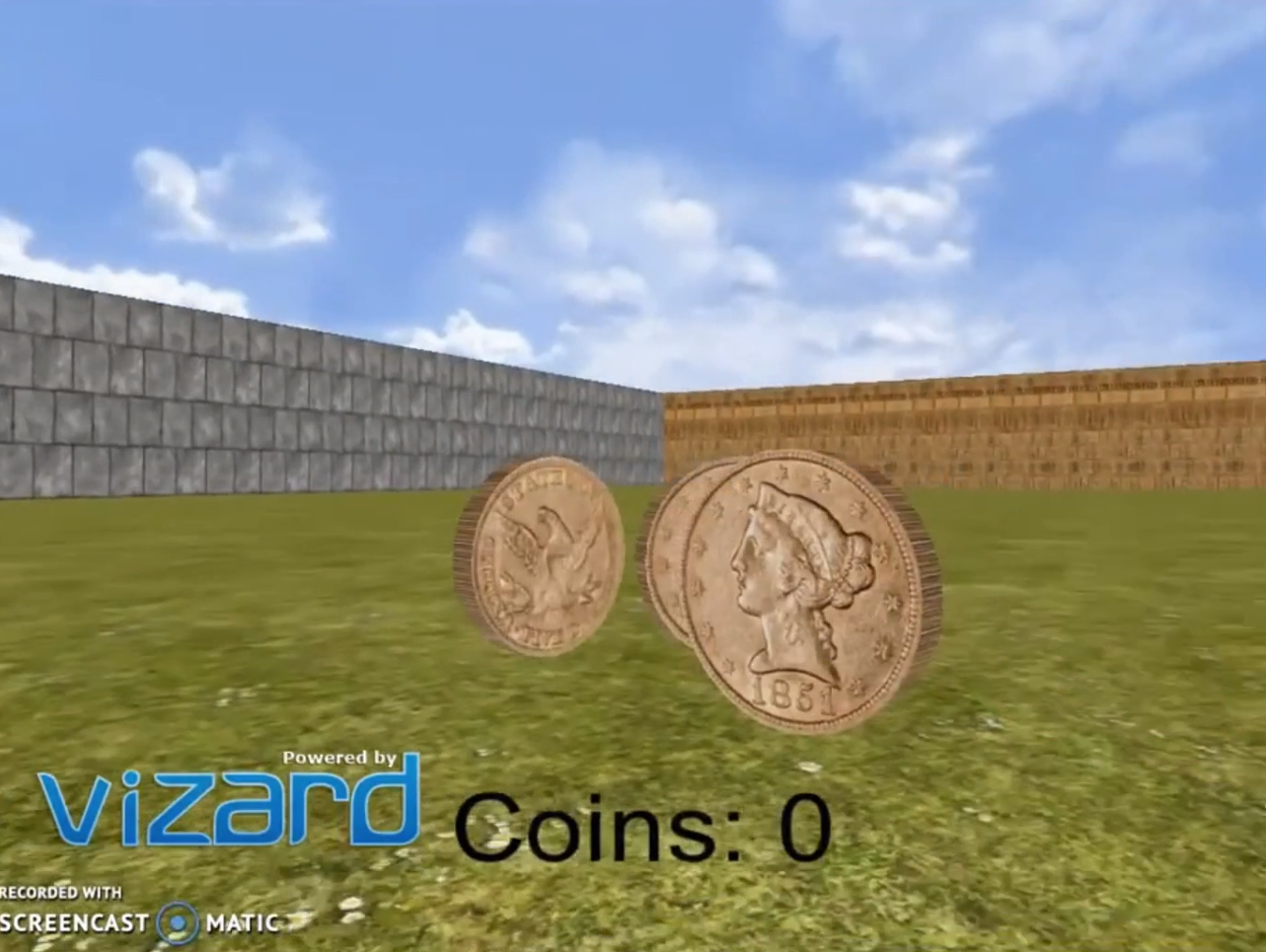}
        \caption{}
        \label{fig:foraging_experiment_view1}
    \end{subfigure}%
    ~
    \begin{subfigure}[t]{0.3\textwidth}
        \centering
        \includegraphics[width=\linewidth]{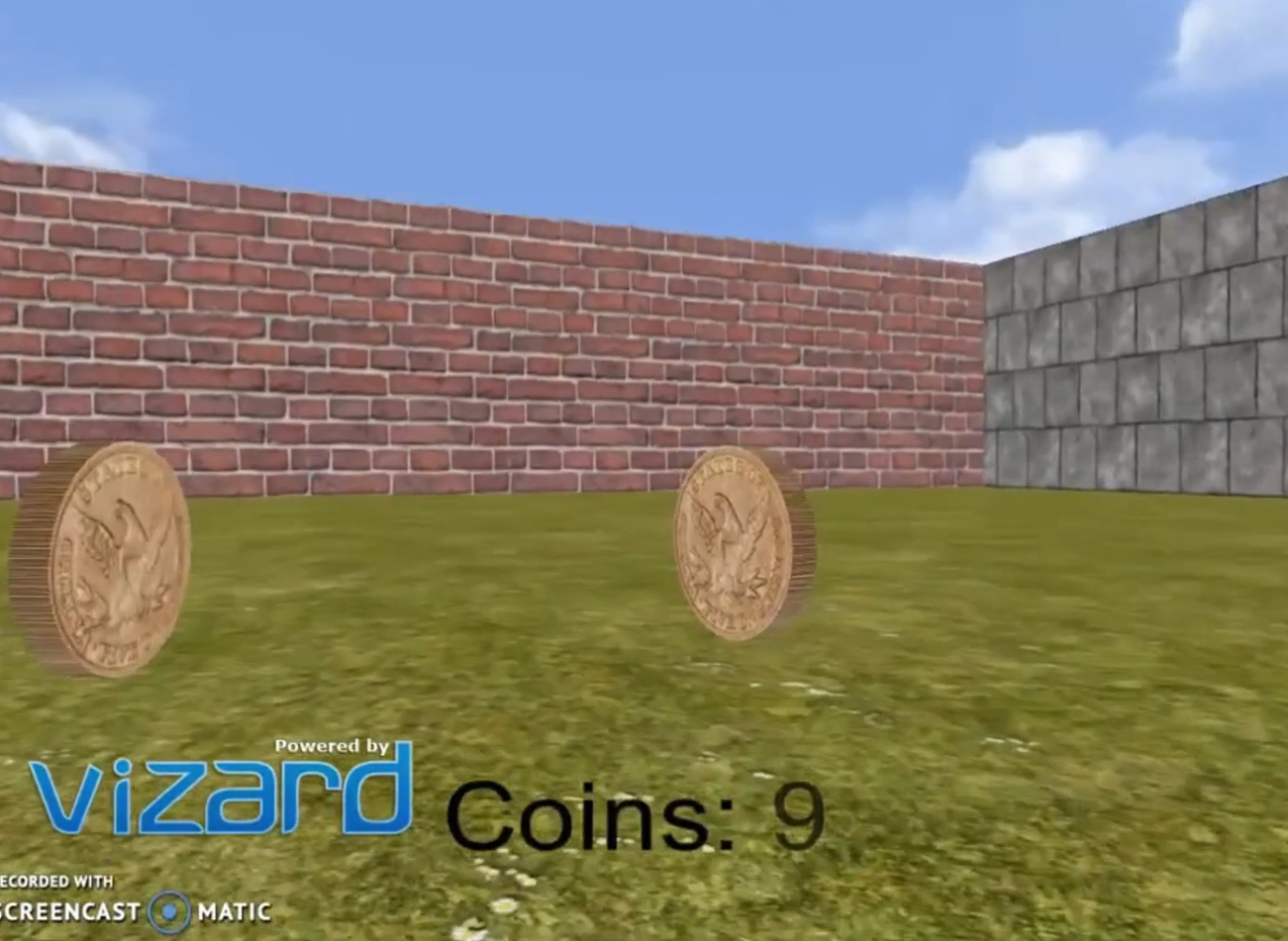}
        \caption{}
        \label{fig:foraging_experiment_view2}
    \end{subfigure}
    ~
    \begin{subfigure}[t]{0.3\textwidth}
        \centering
        \includegraphics[width=0.8\linewidth]{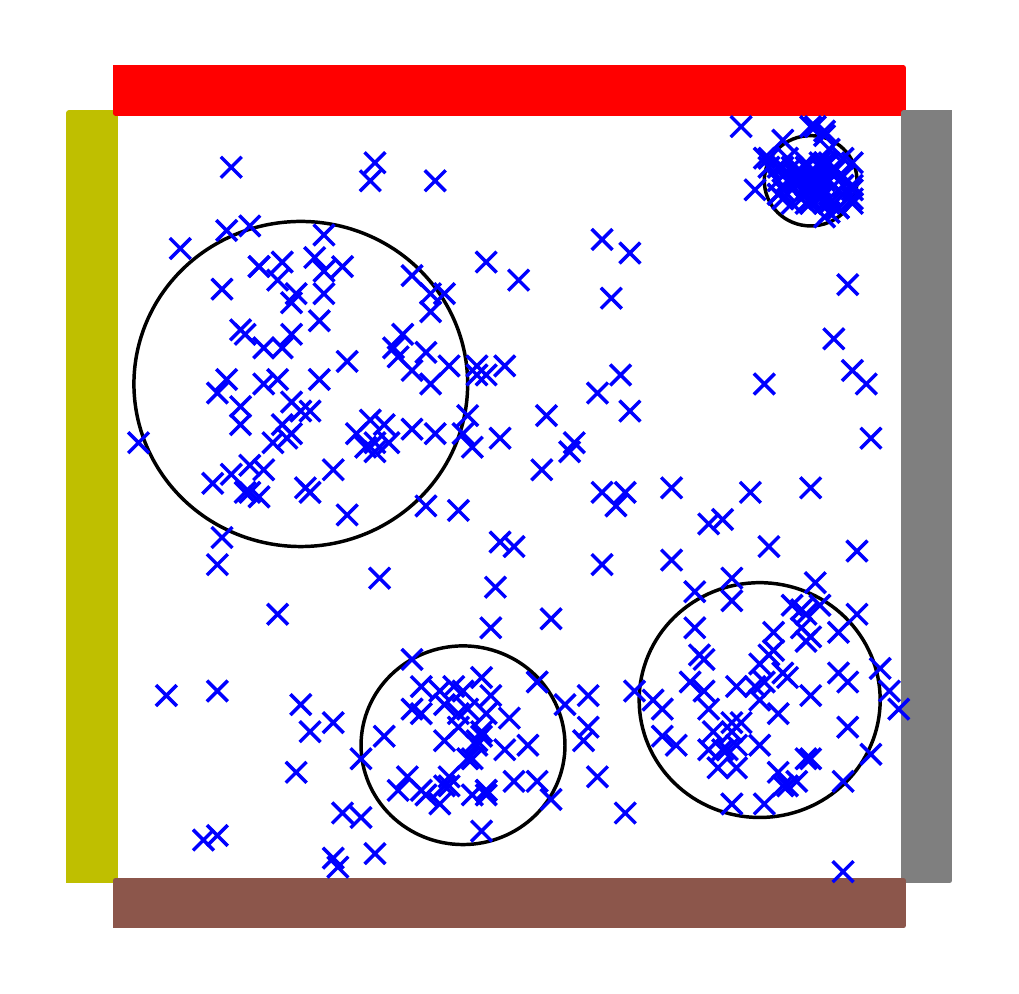}
        \caption{}
        \label{fig:top_view}
    \end{subfigure}
    \caption{Fig.~\ref{fig:foraging_experiment_view1} and \ref{fig:foraging_experiment_view2} are two snapshots of the foraging game in which the forager is about to collect coins. Fig.~\ref{fig:top_view} shows a top view of the environment where the coins are indicated by crosses and the coin clusters are indicated by circles.}
    \label{fig:foraging_experiment}
\end{figure}

\subsection*{Data Availability Statement}
The dataset and the code used to generate our results are freely accessible at our \href{https://github.com/VittorioGiammarino/Learning-from-humans-combining-imitation-and-deep-on-policy-reinforcement-learning-to-accomplish-su}{GitHub repository}.

\subsection*{Experimental Setup}
In the following section, we provide a description of how the human foraging datasets were collected. In the supplementary materials we include the full dataset of $50$ search trajectories. We focus on five participants in the context of a larger study investigating human foraging.\cite{Moore2021virtual} An example of two foraging search trajectories is given in Fig.~\ref{fig:Two_sample_Trajs}. All the experiments have been carried out in accordance with the relevant guidelines and regulations and approved by the Boston University's Institutional Review Board.

\paragraph{Participants:} Participants consisted of male and female, neurologically healthy, English-speaking volunteers between the ages of 18-35 with normal or corrected-to-normal vision. Participants were recruited from Boston University and the surrounding community. Individuals with a history of drug abuse, use of psychoactive medication, neurological or psychiatric disorders, or learning disabilities were excluded. Additionally, participants with a history of motion sickness when watching or playing video games were also excluded. All participants were compensated and gave written informed consent in accordance with Boston University's Institutional Review Board.

\paragraph{Task:} The task consisted of a $160m \times 160m$ virtual “open-field”, i.e. obstacle free, paradigm surrounded by four differently colored and textured walls created using Vizard 6.0, a Python-based virtual reality development platform (Fig.~\ref{fig:foraging_experiment}). 325 coins were distributed throughout the environment, of which 100 were randomly distributed and 225 were distributed according to four different multivariate Gaussian distributions of varying sizes: $75$ according to $\mathcal{N}((60, 75),5^2\bI)$, $40$ according to $\mathcal{N}((-15, -50), 11^2\bI)$, $60$ according to $\mathcal{N}((-50, 30),18^2\bI)$ and $50$ according to $\mathcal{N}((49, -40),13^2\bI)$ (Fig.~\ref{fig:top_view}). Each participant's starting location was randomized at the beginning of each run. Participants could move forward and turn left or right. They could not move backwards. They were instructed to freely explore the environment and collect as many coins as possible but were not told anything about the distribution or total number of coins. They were also able to see a running count of the coins they had collected for each run. After being collected each coin disappears for the remaining duration of the run. Participants performed the foraging task over two consecutive days. On the first day, naive participants were presented with the task on a desktop computer in a behavioral testing room. On the second day, they performed the same task in an MRI scanner. Subjects performed 10 eight-minute runs on Day 1 and 10 eight-minute runs on Day 2. In the desktop condition (Day 1), participants moved using keyboard arrow keys, and in the scanner (Day 2), they moved using a diamond-shaped button box. For our purposes here, we utilize the 80 minutes of behavioral data from Day 2 of the experiment for 5 participants which collected an average of $243.98$ coins each. 

\subsection*{Modeling the Human Decision Process}

\begin{figure}[ht]
    \centering
    \includegraphics[width=11cm,height=5cm]{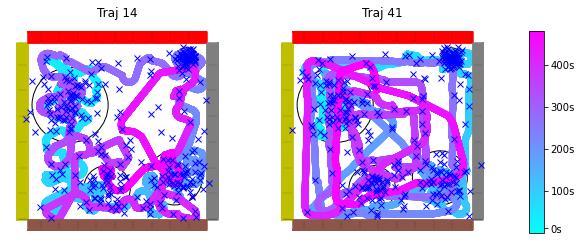}
    \caption{Two samples trajectories collected during the second day of tests. The bar on the right shows the time in seconds. The full set of the $50$ trajectories is available in the supplementary materials.}
    \label{fig:Two_sample_Trajs}
\end{figure}

In this section we describe the human modeling step and how the data are processed to make them suitable for IL and RL.

We consider an infinite-horizon discounted Markov Decision Process (MDP) defined by the tuple $(\mathcal{S}, \mathcal{A}, P, r, D, \gamma)$ where $\mathcal{S}$ is the finite set of states and $\mathcal{A}$ is the finite set of actions. $P:\mathcal{S}\times \mathcal{A} \rightarrow \Delta_{\mathcal{S}}$ is the transition probability function and $\Delta_{\mathcal{S}}$ denotes the space of probability distributions over $\mathcal{S}$. The function $r:\mathcal{S}\times \mathcal{A} \rightarrow \mathbb{R}$ maps rewards to state-action pairs. $D\in\Delta_{\mathcal{S}}$ is the initial state distribution and $\gamma \in [0,1)$ the discount factor. The decision agent is modeled as a stationary policy $\pi:\mathcal{S}\rightarrow\Delta_{\mathcal{A}}$, where $\pi(a|s)$ is the probability of taking action $a$ in state $s$. When a deterministic policy is required we simply take $a = \arg\max_{a'} \pi(a'|s)$. For simplicity, we will always write $a \sim \pi(\cdot|s)$ and according to which algorithm we are referring to it will be clear whether $\pi$ is stochastic or deterministic. We parameterize $\pi$ using a neural network with parameters $\bm{\theta} \in \varTheta \subset \mathbb{R}^k$ and we write $\pi_{\bm{\theta}}$. 

Given an MDP, we consider the human participants taking into account both egocentric and allocentric strategies when navigating. \cite{alexander2020egocentric, feigenbaum2004allocentric} We define the state vector as $\bm{s} = \{x,y,\psi,\chi\}$, where $x,y$ are coordinates with respect to a frame fixed to the environment and represent the allocentric capacities of the agent, i.e., the ability to approximate its current position within the environment. Instead, $\psi$ and $\chi$ are two categorical variables that describe the human egocentric behavior: the first tells the agent whether it can see a coin or not in its vicinity, $\psi \in \{$ see coin, no coins $\}$, the second describes the "greedy" direction, i.e., the direction of the closest coin the agent has in its view, 
$\chi \in \{\text{east},$ northeast, north, northwest, west, southwest, south, southeast, \text{no coins}$\}$. 

The artificial agent perceives the state and takes an action to interact with the environment. For computational reasons we discretize the $x,y$ coordinates on a fine grid of $1m \times 1m$ and define the action space as $a \in \{\text{east},$ northeast, north, northwest, west, southwest, south, $\text{southeast}\}$. The transition to the next state always occurs deterministically in the direction of the action $a$ taken by the agent. As convention in Fig.~\ref{fig:Two_sample_Trajs}, north means going from bottom to top, south from top to bottom, east is left to right and west vice versa. The categorical state $\psi$ stays $0$ all the time unless there is a coin in a radius of $8m$ distance, when $\psi$ turns $1$ then also $\chi$ turns from "no coins" to one of the other directions. This is aligned with the original experiment where each coin pops-up when the human is at $8m$ from it. Finally, the rewards are simply represented by the coins in the environment where $r(\bm{s},a)=1$ for each coin collected. As in the original experiment, the agents automatically collect the reward once at $3m$ and $D$ is a uniform distribution over $\mathcal{S}$. 

\subsection*{Imitation Learning}

Given a task and an agent performing the task, IL infers the underlying agent distribution via a set of an agent's demonstrations (state-action samples). Assuming the agent's behavior is parameterized by a NN with optimal parameters $\bm{\theta}^*$, we refer to the process of estimating $\bm{\theta}^*$ through a finite sequence of agent's demonstrations $\tau = (\bm{s}_{0:T},a_{0:T})$ with $2 \leq T < \infty$ as IL. One way to formulate this problem is through maximum likelihood estimation:
\begin{equation}
    \max_{\bm{\theta}}  \mathcal{L}(\bm{\theta}),
    \label{eq:IL_max_likelihood}
\end{equation}
where $\mathcal{L}(\bm{\theta})$ denotes the log-likelihood and is equivalent to the logarithm of the joint probability of generating the expert demonstrations $\tau = \{\bm{s}_0,a_0,\bm{s}_1,a_1,\dots,\bm{s}_T,a_T\}$, i.e.,
\begin{equation}
    \mathcal{L}(\bm{\theta}) = \log \mathbb{P}^{\bm{\theta}}_D(\tau).
    \label{eq:log_likelihood}
\end{equation}
$\mathbb{P}^{\bm{\theta}}_D(\tau)$ in \eqref{eq:log_likelihood} is defined as 
\begin{align}
    \begin{split}
        &\mathbb{P}^{\bm{\theta}}_{D}(\tau) = 
        D(\bm{s}_0)\bigg[\prod_{t=0}^{T} \pi_{\bm{\theta}}(a_t|\bm{s}_t) \bigg]\bigg[\prod_{t=0}^{T-1}P(\bm{s}_{t+1}|\bm{s}_t,a_t)\bigg].
    \end{split}
    \label{eq:Preliminary_joint_distribution}
\end{align}
Computing the logarithm of \eqref{eq:Preliminary_joint_distribution} and neglecting the elements not parameterized by $\bm{\theta}$ we obtain the following maximization problem
\begin{equation}
    \max_{\bm{\theta}}  \sum_{t=0}^{T}\log \big(\pi_{\bm{\theta}}(a_t|\bm{s}_t) \big).
    \label{eq:IL_max_prob}
\end{equation}
Solving the maximization problem in Eq.~\eqref{eq:IL_max_prob} is the main objective of our IL step. 

\subsection*{Reinforcement Learning}

After defining a model, collecting the data and performing the imitation step, our final goal is to further refine the imitated policies using RL. In RL, the artificial agents are allowed to experience the task themselves and receive a reinforcement according to the reward function $r(\bm{s}_t,a_t)$. Mathematically, the goal is to find the policy parameters $\bm{\theta}$ which maximize the expected total discounted reward $J(\bm{\theta})=\mathbb{E}_{\tau}[\sum_{t=0}^{\infty}\gamma^t r(\bm{s}_t,a_t)]$, where, as previously, $\tau = (\bm{s}_0,a_0,\bm{s}_1,a_1,\dots)$ is sampled according to $\bm{s}_0 \sim D$, $a_t\sim\pi_{\bm{\theta}}(\cdot|\bm{s}_t)$ and $\bm{s}_{t+1}\sim P(\cdot|\bm{s}_t,a_t)$. Our focus is on model-free RL methods in which the artificial agent does not know the transition probability function $P(\cdot|\bm{s}_t,a_t)$, and it can only explore the environment and experience rewards. Among these types of algorithms, we can distinguish two main groups: $(i)$ algorithms that update the current policy following the agent's generated trajectories according to this policy, also known as {\em on-policy} algorithms, and $(ii)$ algorithms that update the current policy using experience from multiple policies used previously, known as {\em off-policy}. We provide a more thorough introduction on this difference in the supplementary materials. 

State-of-art on-policy algorithms include Trust Region Policy Optimization (TRPO),\cite{schulman2015trust} and some robust variants such as Uncertainty Aware TRPO (UATRPO),\cite{queeney2021uncertainty}, and 
Proximal Policy Optimization (PPO).\cite{schulman2015trust} Whereas, off-policy methods include the Soft-Actor Critic (SAC) and Twin Delayed Deep Deterministic Policy Gradient (TD3).\cite{haarnoja2018soft, fujimoto2018addressing} All the mentioned algorithms are tested in our experiments and more details about the NNs design are available in the supplementary materials. 

\section*{Results}
In this section, we present our results and describe all the steps that lead to our design. All the code and data to replicate the experiments are freely accessible at our \href{https://github.com/VittorioGiammarino/Learning-from-humans-combining-imitation-and-deep-on-policy-reinforcement-learning-to-accomplish-su}{GitHub repository} (https://github.com/VittorioGiammarino/Learning-from-humans-combining-imitation-and-deep-on-policy-reinforcement-learning-to-accomplish-su). An overview of the NNs used to parameterize $\pi_{\bm{\theta}}$ and all the hyperparameters used for each of the IL and RL algorithms are in the supplementary materials. 

\subsection*{Pre-processing}
Our first step is to collect and process the $50$ trajectories, of $8$ minutes each, recorded on the second day of tests. Each $8$-minute trajectory consists of $28973$ data points, on average, which means we collect a data point every $0.017s$, where the data points are the human agent's coordinates with respect to the fixed environment frame. Note that it is possible that a human agent does not move for a few seconds, for example, and then makes many rapid decisions about where to explore in the next milliseconds following this stationary period. Therefore, the first "few seconds" could be aggregated in a single data point while the next "milliseconds" would require more than a single point. As a result, we aggregate the data points considering the discretization of the $(x,y)$ coordinates. After that, we go over each of the trajectories and determine the human decisions (i.e., for each aggregated state, the direction of the human's next movement). We cast each human decision for each trajectory in the pre-determined action space $\mathcal{A}$ and construct in this way our state-action pairs (i.e., actions $a$ taken at state $\bm{s}$). This process allows us to reduce the average length of human trajectories from $28973$ to $3464$ data points without losing key information. Note that this processing is an expensive but necessary step for reducing the computational burden and enabling learning. Future research will focus on how to automate this step and developing methods which can handle learning from raw data.

\subsection*{Imitation Learning}
\begin{figure}[ht]
    \centering
    \includegraphics[width=12cm,height=3.5cm]{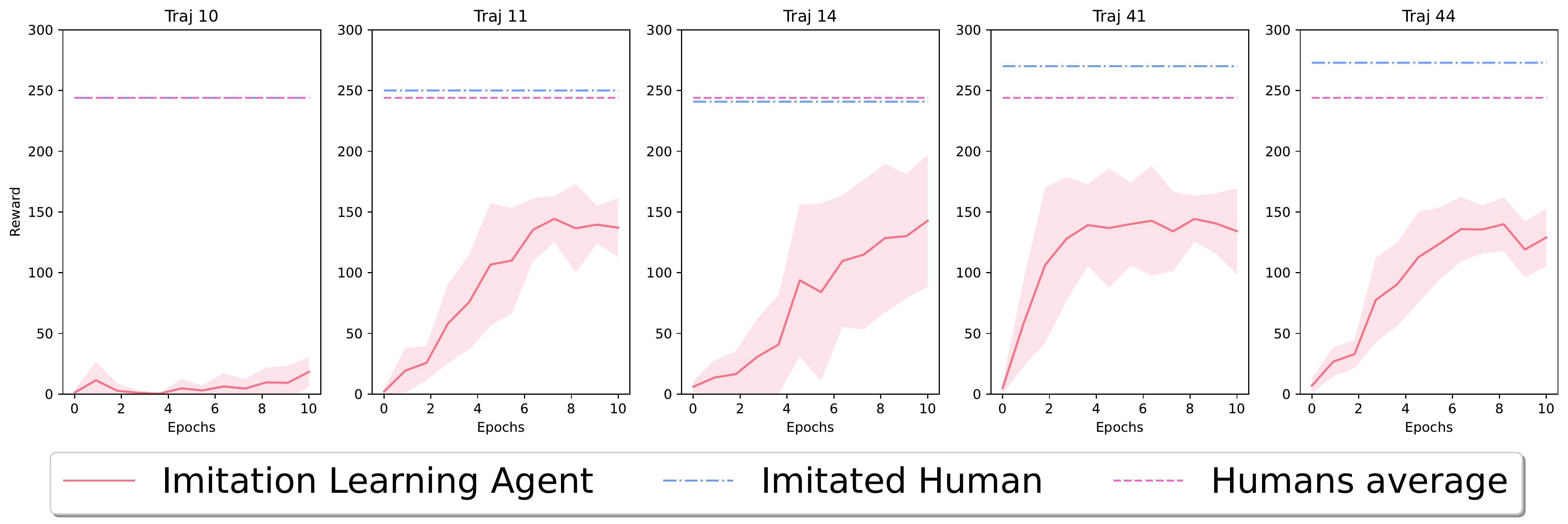}
    \caption{The results from the IL step for $5$ different human trajectories are illustrated. For each human trajectory we solve the IL problem for $8$ different seeds. For each seed, after every epoch we evaluate the performance of the learnt policy for $10$ trials, each consisting of $3464$ steps. The reported results for the "Imitation Learning Agent" show the reward averaged over $10$ trials and $8$ seeds; the shaded area shows the standard deviation over the seeds. The performance of the human trajectory used as data set for the imitation is labelled as "Imitated Human". "Humans average" is the average performance of the $50$ human trajectories.}
    \label{fig:HIL_comparison_selected}
\end{figure}

We perform IL on each human trajectory individually rather than considering a single data set with all the trajectories. This is due to two main factors: first, as Fig.~\ref{fig:Two_sample_Trajs} shows, each human trajectory covers the majority of the environment; hence, each trajectory is per se informative enough about the task. Second, the trajectories are noisy and a single aggregated data set turns to be too noisy to be beneficial. The results of the IL step and all the details on the evaluation are illustrated, for $5$ humans' trajectories, in Fig~\ref{fig:HIL_comparison_selected}. In summary, we achieve good learning performance for several trajectories but not enough to match the human participants. A figure showing the IL performance for all the $50$ trajectories is available in the supplementary materials.

\subsection*{Reinforcement Learning}

\begin{figure}[ht]
    \centering
    \includegraphics[width=11cm,height=4cm]{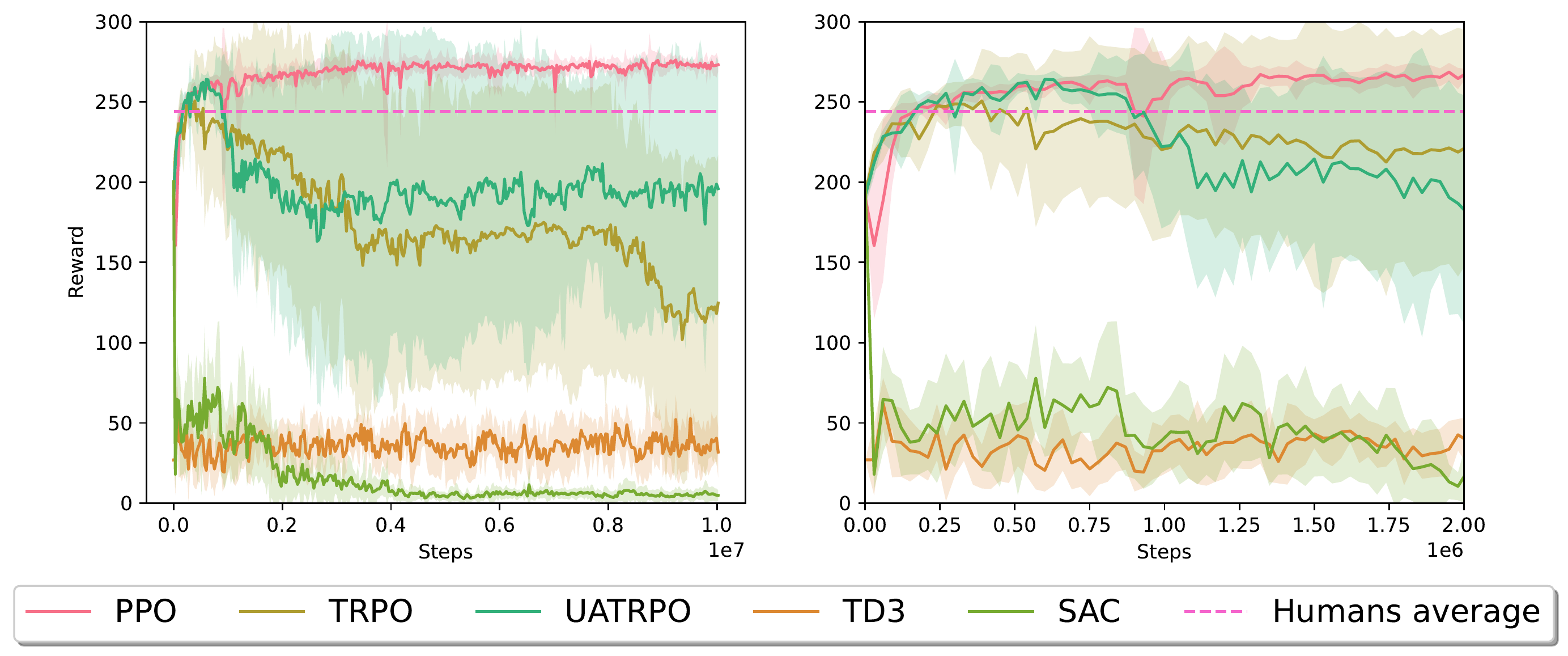}
    \caption{The results for the RL algorithms initialized with the same policy are illustrated. For each algorithm we run the learning process for $10^7$ steps and $8$ different random seeds. For each seed, after every $30,000$ steps we evaluate the performance of the learnt policy for $10$ trials each of $3464$ steps. The reported results show the reward averaged over $10$ trials and $8$ seeds, the shaded area shows the standard deviation over seeds.}
    \label{fig:RL_algortihms_comparison}
\end{figure}

\begin{figure}[ht]
    \centering
    \includegraphics[width=12cm,height=3.5cm]{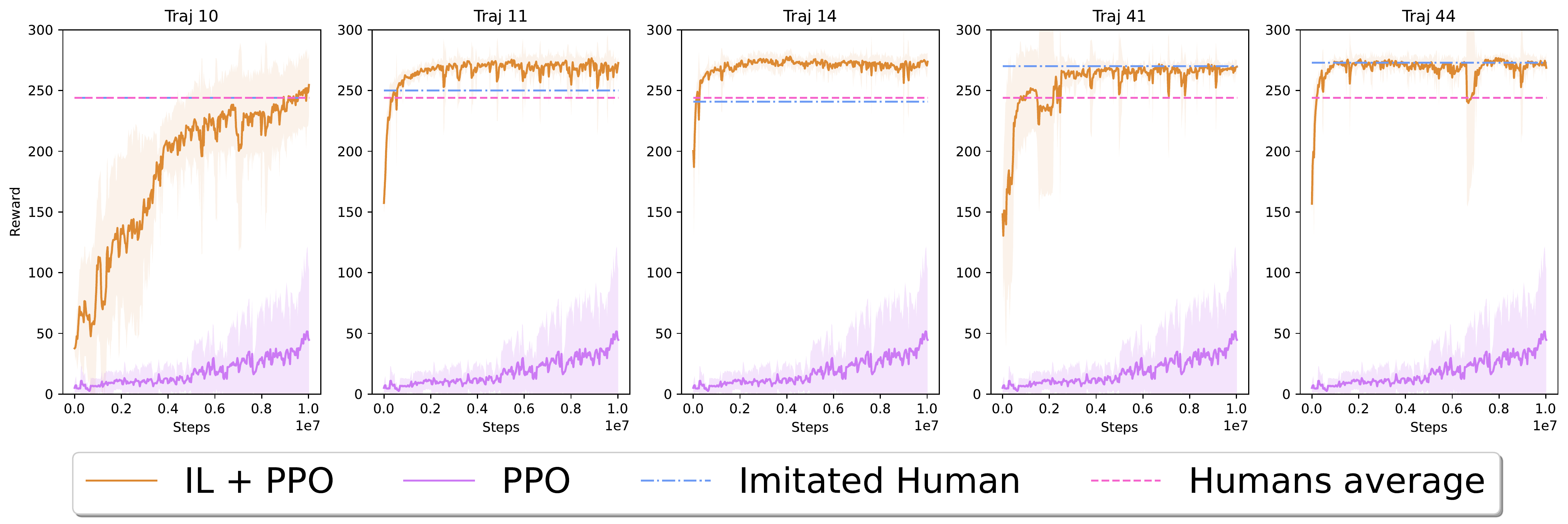}
    \caption{The results of the IL+PPO method compared with the PPO-only alternative, the average human performance and the performance of the imitated expert are illustrated for the trajectories of Fig.~\ref{fig:HIL_comparison_selected}. The experiment design and the reported results follow the same criterion as in Fig.~\ref{fig:RL_algortihms_comparison}.}
    \label{fig:PPO_comparison_selected}
\end{figure}

We consider the $50$ policies learnt from the $50$ human trajectories during the IL step. We refine these policies using RL. We design the experiment as follows: 
\begin{enumerate}
    \item First, in order to determine which RL algorithm is more suitable for our goal, we take the same single policy learnt during the IL step and use it as initialization of each of the RL algorithms.
    \item Given the state-space dimension, we consider $10^7$ steps as a reasonable amount of steps for performing RL.
    \item As in the IL step, for each RL algorithm we run the learning process for $8$ random seeds.
    \item During the learning process, we evaluate the policy learnt every $30,000$ steps on $10$ trials of $3464$ steps each. We report averaged results over the $10$ trials and $8$ seeds. The shaded area in Fig.~\ref{fig:RL_algortihms_comparison} and \ref{fig:PPO_comparison_selected} shows the standard deviation over seeds.
    \item After determining the most suitable RL algorithm, we rerun the whole experiment for $50$ times in which each of the $50$ policies learnt during the IL step is used as initialization of the selected RL algorithm.
\end{enumerate}

Fig.~\ref{fig:RL_algortihms_comparison} compares the various RL algorithms. PPO outperforms all other methods. Broadly speaking, on-policy algorithms, i.e., PPO, TRPO, and UATRPO, learn more effectively from a pre-initialized policy with respect to the off-policy algorithms TD3 and SAC. Refer to the supplementary materials for more details.

Consequently, we proceed by combining IL together with PPO and compare it with the PPO-only alternative. Fig.~\ref{fig:PPO_comparison_selected} illustrates the final results for the same trajectories of Fig.~\ref{fig:HIL_comparison_selected} and a figure showing this final result for all the $50$ trajectories is available in the supplementary materials. In summary, IL followed by PPO (IL+PPO) outperforms the average human performance and its imitated expert, $39$ ($78\%$) and $31$ ($62\%$) times, respectively, over the $50$ human trajectories. On the other hand, the PPO-only alternative cannot get close to these results in $10^7$ steps. Table~\ref{tab:method_vs_humans_results} summarizes the comparison between humans and IL+PPO policies with respect to the total amount of collectable rewards. 

Note that, Fig.~\ref{fig:PPO_comparison_selected} provides interesting insights on why pre-training with IL makes sense in foraging tasks with sparse rewards. We observe that, in addition to a different initial performance, the IL+PPO and the PPO-only agents show really different exploration strategies which lead to a different reward convergence rate (the difference in rates is clearly visible in Fig.~\ref{fig:PPO_comparison_selected}). The human-inspired exploration strategy of IL+PPO represents the main strength of the method in this setup and the main source of difference with the PPO-only agents. 

\begin{table}[h!]
\centering
\caption{A summary of the comparison between IL+PPO and human performance in the original experiment. The table shows the percentage out of $50$ learnt policies, for both the IL+PPO and the human agents, where the agent collected at least a certain percentage of rewards. As an example, the table is showing that, for the IL+PPO method, $19$ policies ($38\%$) can collect at least $260$ coins ($>80\%$). We compare these results with the human performance where each trajectory is considered as a single human policy. Note that, the IL+PPO policies perform better than humans on average but cannot do better than the best participants.}
\label{tab:method_vs_humans_results}
\begin{tabular}{l c c c c c c c c c}\toprule
& & \multicolumn{8}{c}{\textbf{Performance Lower Bound}}\\
& & \multicolumn{2}{c}{ $>70\%$} & \multicolumn{2}{c}{$>80\%$} & \multicolumn{2}{c}{$>85\%$} & \multicolumn{2}{c}{$>90\%$} \\
\cmidrule(lr){3-10}
\multicolumn{2}{c}{\textbf{IL+PPO}} & \multicolumn{2}{c}{$88\%$} & \multicolumn{2}{c}{$38\%$} & \multicolumn{2}{c}{$0\%$} & \multicolumn{2}{c}{$0\%$} \\
\multicolumn{2}{c}{\textbf{Humans}} & \multicolumn{2}{c}{$80\%$} & \multicolumn{2}{c}{$22\%$} & \multicolumn{2}{c}{$10\%$} & \multicolumn{2}{c}{$0\%$} \\
\bottomrule
\end{tabular}
\end{table}

\subsection*{Robustness to reward distribution shift and the importance of egocentric representations}
\begin{figure}[ht]
    \centering
    \begin{subfigure}[t]{0.45\textwidth}
        \centering
        \includegraphics[width=3.5cm,height=3.5cm]{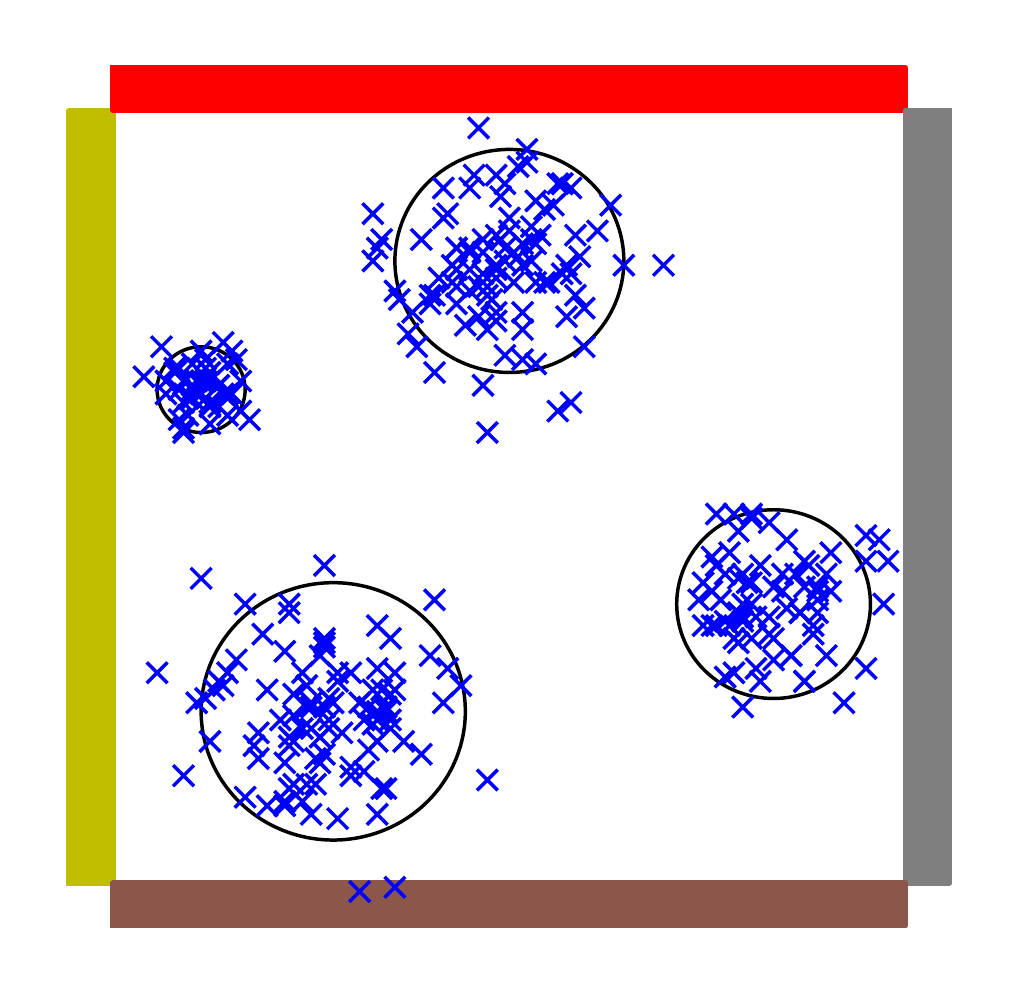}
        \caption{New rewards distribution.}
        \label{fig:top_view_ADV}
    \end{subfigure}
    ~
    \begin{subfigure}[t]{0.45\textwidth}
        \centering
        \includegraphics[width=3.5cm,height=3.5cm]{Figures/Environment_top_view.pdf}
        \caption{Old rewards distribution.}
        \label{fig:top_view_comparison}
    \end{subfigure}
    \caption{Fig.~\ref{fig:top_view_ADV} shows the new reward distribution that was not previously seen by any of the artificial agents. We include the previous reward distribution in Fig.~\ref{fig:top_view_comparison} to facilitate the comparison.}
    \label{fig:robustness_experiment}
\end{figure}

\begin{table}[h!]
\centering
\caption{A summary of the results for the rewards distribution shift experiment. The table shows the fraction out of $50$ policies, for each initialization method, where after learning for $2\times10^6$ steps, the agent is able to collect at least a certain percentage of rewards. As an example, the table is showing that, for the IL+PPO initialization, $46$ policies ($92\%$) can collect at least $270$ coins ($>80\%$) in this new scenario after learning for only $2\times10^6$ steps. }
\label{tab:robustness_results}
\begin{tabular}{l c c c c c c c c c}\toprule
& & \multicolumn{8}{c}{\textbf{Performance Lower Bound}}\\
& & \multicolumn{2}{c}{ $>70\%$} & \multicolumn{2}{c}{$>80\%$} & \multicolumn{2}{c}{$>90\%$} & \multicolumn{2}{c}{$>95\%$} \\
\cmidrule(lr){3-10}
\multicolumn{2}{c}{\textbf{IL-only initialization}} & \multicolumn{2}{c}{$28\%$} & \multicolumn{2}{c}{$18\%$} & \multicolumn{2}{c}{$0\%$} & \multicolumn{2}{c}{$0\%$} \\
\multicolumn{2}{c}{\textbf{IL+PPO initialization}} & \multicolumn{2}{c}{$98\%$} & \multicolumn{2}{c}{$92\%$} & \multicolumn{2}{c}{$18\%$} & \multicolumn{2}{c}{$0\%$} \\
\bottomrule
\end{tabular}
\end{table}

In this section, we test the learnt policies for robustness to a reward distribution shift. The motivation is to explore how quickly the artificial agents grasp changes in the environment and adapt to these changes. As in the original experiment, $325$ coins are placed across the environment; however, this time according to the new distribution in Fig.~\ref{fig:top_view_ADV}, we include the original coins distribution in Fig.~\ref{fig:top_view_comparison} to facilitate the comparison with Fig.~\ref{fig:top_view_ADV}. Specifically, $50$ coins are distributed according to $\mathcal{N}((-70, 30),5^2\bI)$, $75$ according to $\mathcal{N}((60, -20), 11^2\bI)$, $100$ according to $\mathcal{N}((-40, 45),15^2\bI)$ and $100$ according to $\mathcal{N}((0, 60),13^2\bI)$.

We design the experiment similarly to the RL study and the IL+PPO experiments in Fig.~\ref{fig:RL_algortihms_comparison} and Fig.~\ref{fig:PPO_comparison_selected}. Overall, we run, for $8$ different random seeds, $100$ learning experiments of $2\times10^6$ steps each, where in the first $50$ we initialize using the policies learnt with only IL (Fig.~\ref{fig:HIL_comparison_selected}), while, in the second $50$, we initialize using the policies learnt by IL+PPO (Fig.~\ref{fig:PPO_comparison_selected}). The results are summarized in Table~\ref{tab:robustness_results} and show that the policies learnt using both IL+PPO generalize well to novel reward distributions. The figures showing the detailed $100$ experiments are available in the supplementary materials.

In order to produce these results, we conclude that, given the state vector representation as $\bm{s} = \{x,y,\psi,\chi\}$, the RL agents and their exploration strategies must heavily rely on egocentric information, i.e., the variables $\psi$ and $\chi$. This would explain the algorithm performance in the novel reward environment in Fig.~\ref{fig:top_view_ADV}, where the previously learnt allocentric representation is no longer informative. On the other hand, a strategy based on egocentric exploration which facilitates the generation of a new allocentric representation of the environment would explain the results in Table~\ref{tab:robustness_results}. In other words, we suggest that our IL+PPO algorithms exhibit coding of behavioral variables analogous to the observation in animals,\cite{alexander2020egocentric} where electrophysiological recording during foraging strategies indicate neural coding in both egocentric and allocentric coordinate frames. 

To demonstrate the veracity of this claim we rerun the entire set of experiments, which includes IL as in Fig.~\ref{fig:HIL_comparison_selected} and IL+PPO as in Fig.~\ref{fig:PPO_comparison_selected}, but this time only providing allocentric information to the artificial agents. In other words, we reduce the state vector from $\bm{s} = \{x,y,\psi,\chi\}$ to $\bm{s} = \{x,y\}$. The results for a selected number of human trajectories are summarized in Fig~\ref{fig:allocentric_only_experiment} and for the entire set of trajectories in the supplementary materials. The final results show that agents trained with the full state $\bm{s} = \{x,y,\psi,\chi\}$ outperform agents trained with the allocentric only state $\bm{s} = \{x,y\}$ $74\%$ of the times for IL ($37$ out of $50$) and $100\%$ of the times for IL+PPO. We conclude that our learnt policies heavily rely on egocentric data and that the absence of such information compromises to a large extent the learning performance as illustrated in Fig~\ref{fig:allocentric_only_experiment}.
\begin{figure}[ht]
    \centering
    \begin{subfigure}[t]{\textwidth}
        \centering
        \includegraphics[width=12cm,height=3.5cm]{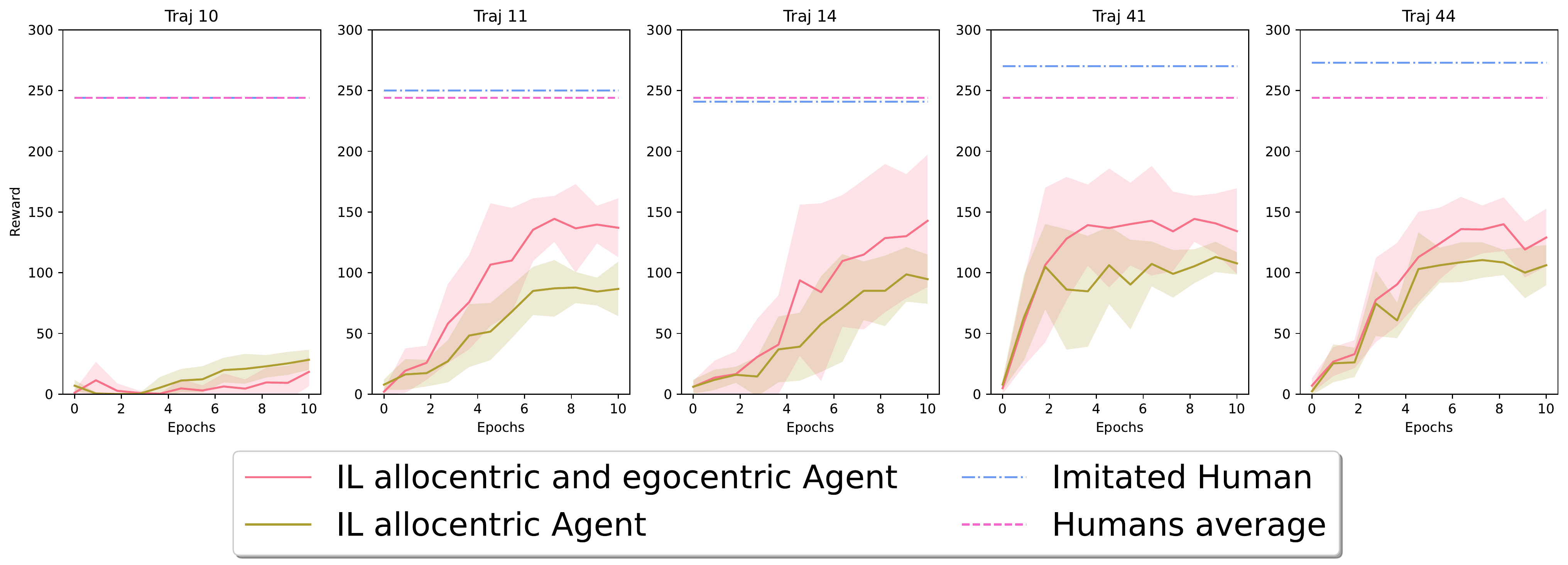}
    \end{subfigure}
    ~
    \begin{subfigure}[t]{\textwidth}
        \centering
        \includegraphics[width=12cm,height=3.5cm]{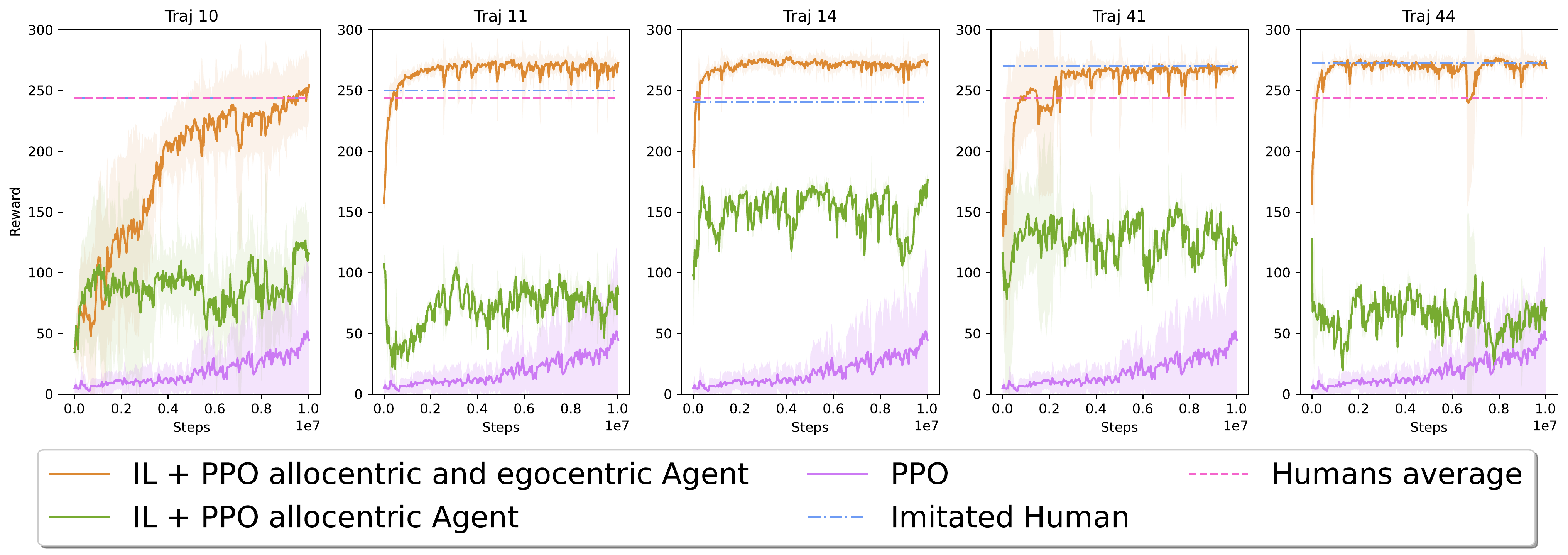}
    \end{subfigure}
    \caption{Performance comparison, for selected trajectories, among agents trained with full state including both egocentric and allocentric information $\bm{s} = \{x,y,\psi,\chi\}$ versus an allocentric only state $\bm{s} = \{x,y\}$. The comparison is done for both IL (upper figure) and IL+PPO (lower figure). For completeness, we also show the PPO agent without IL initialization performance (PPO allocentric and egocentric Agent).} %Please put the description of the color of lines into the figure caption, instead of the figure, and make sure to label IL+PPO as egocentric+allocentric. This is better than trying to show the colors on the figure itself because that will run into problems with size of script on figure.
    \label{fig:allocentric_only_experiment}
\end{figure}

\section*{Discussion}

In this paper, $50$ human navigation trajectories were collected in a virtual open-field environment. We extracted a navigation control policy from each of these trajectories and introduced an MDP setting to capture the navigational human decision making. We learned policies consistent with the experimental data using imitation learning based on log-likelihood maximization for each of the trajectories. 

After obtaining a control policy for each trajectory, we used all of them as a starting point for RL, seeking to find policies that can efficiently outperform the human participants in the same experimental setting. We tested state-of-art on-policy (PPO, TRPO, UATRPO) and off-policy (TD3, SAC) algorithms. We explained more extensively how these two categories differ in the supplementary materials. Briefly, the main element of difference lies in the data used to update the policy network $\pi_{\bm{\theta}}$ and in how we  compute and approximate the critic network. Off-policy algorithms are usually faster to converge but introduce a large bias in the critic estimate, which results in more oscillatory learning which often jeopardizes the IL initialization. On the other hand, the tested on-policy algorithms are more conservative, and the optimization step is constrained so not to diverge too much from the current policy $\pi_{\bm{\theta}}$. This results in a slower but more steady improvement of performance. Our preference towards PPO with respect to the other on-policy algorithms is the result of empirical experiments which corroborates other well-known empirical studies on the matter.\cite{andrychowicz2020matters, engstrom2020implementation} 

Finally, we examined the sensitivity of the IL+PPO and IL-only policies to a different reward distribution and investigated to what extent our artificial agents rely on egocentric information.
The final results showed that learning only from data is not enough to match human performance and does not lead to robustness over the reward distribution (Fig.~\ref{fig:HIL_comparison_selected} and Table~\ref{tab:method_vs_humans_results}). On the other hand, IL followed by PPO (IL+PPO) showed impressive results in the original experiment and it led to good generalization of the task (Fig.~\ref{fig:PPO_comparison_selected} and Table~\ref{tab:robustness_results}). Further, we showed that such results are associated with the use of egocentric information, which are crucial in enhancing learning performance both in the IL and the IL+RL setting when compared with the use of allocentric information alone. 

In summary, we have developed a method to learn bio-inspired policies from human navigation data, which can be further refined to achieve human-level performance. This approach to modeling human navigational policies can be of great utility for aerial and ground unmanned navigation tasks including scientific exploration and search and rescue operations.

\section*{Acknowledgments}
Research was partially supported by the NSF under grants IIS-1914792, DMS-1664644, CNS-1645681 and NSF-1829398, by the ONR under grants N00014-19-1-2571 and N00014-21-1-2844, by the ONR DURIP under grants N00014-17-1-2304, by the NIH under grants R01 GM135930 and UL54 TR004130, and by the Boston University Kilachand Fund for Integrated Life Science and Engineering.   

\bibliography{sample.bib}

\begin{thebibliography}{10}
\urlstyle{rm}
\expandafter\ifx\csname url\endcsname\relax
  \def\url#1{\texttt{#1}}\fi
\expandafter\ifx\csname urlprefix\endcsname\relax\def\urlprefix{URL }\fi
\expandafter\ifx\csname doiprefix\endcsname\relax\def\doiprefix{DOI: }\fi
\providecommand{\bibinfo}[2]{#2}
\providecommand{\eprint}[2][]{\url{#2}}

\bibitem{walker2012explaining}
\bibinfo{author}{Walker, C.~M.}, \bibinfo{author}{Williams, J.~J.},
  \bibinfo{author}{Lombrozo, T.} \& \bibinfo{author}{Gopnik, A.}
\newblock \bibinfo{title}{Explaining influences children's reliance on evidence
  and prior knowledge in causal induction}.
\newblock In \emph{\bibinfo{booktitle}{Proceedings of the Annual Meeting of the
  Cognitive Science Society}}, vol.~\bibinfo{volume}{34}
  (\bibinfo{year}{2012}).

\bibitem{gopnik2015younger}
\bibinfo{author}{Gopnik, A.}, \bibinfo{author}{Griffiths, T.~L.} \&
  \bibinfo{author}{Lucas, C.~G.}
\newblock \bibinfo{journal}{\bibinfo{title}{When younger learners can be better
  (or at least more open-minded) than older ones}}.
\newblock {\emph{\JournalTitle{Current Directions in Psychological Science}}}
  \textbf{\bibinfo{volume}{24}}, \bibinfo{pages}{87--92}
  (\bibinfo{year}{2015}).

\bibitem{goddu2020transformations}
\bibinfo{author}{Goddu, M.~K.}, \bibinfo{author}{Lombrozo, T.} \&
  \bibinfo{author}{Gopnik, A.}
\newblock \bibinfo{journal}{\bibinfo{title}{Transformations and transfer:
  Preschool children understand abstract relations and reason analogically in a
  causal task}}.
\newblock {\emph{\JournalTitle{Child Development}}}
  \textbf{\bibinfo{volume}{91}}, \bibinfo{pages}{1898--1915}
  (\bibinfo{year}{2020}).

\bibitem{ruggeri2021toddlers}
\bibinfo{author}{Ruggeri, A.}, \bibinfo{author}{Pelz, M.},
  \bibinfo{author}{Gopnik, A.} \& \bibinfo{author}{Schulz, E.}
\newblock \bibinfo{journal}{\bibinfo{title}{Toddlers search longer when there
  is more information to be gained}}.
\newblock {\emph{\JournalTitle{PsyArXiv preprint}}}  (\bibinfo{year}{2021}).

\bibitem{silver2017mastering}
\bibinfo{author}{Silver, D.} \emph{et~al.}
\newblock \bibinfo{journal}{\bibinfo{title}{Mastering the game of go without
  human knowledge}}.
\newblock {\emph{\JournalTitle{Nature}}} \textbf{\bibinfo{volume}{550}},
  \bibinfo{pages}{354--359} (\bibinfo{year}{2017}).

\bibitem{botvinick2019reinforcement}
\bibinfo{author}{Botvinick, M.} \emph{et~al.}
\newblock \bibinfo{journal}{\bibinfo{title}{Reinforcement learning, fast and
  slow}}.
\newblock {\emph{\JournalTitle{Trends in Cognitive Sciences}}}
  \textbf{\bibinfo{volume}{23}}, \bibinfo{pages}{408--422}
  (\bibinfo{year}{2019}).

\bibitem{offerman1998learning}
\bibinfo{author}{Offerman, T.} \& \bibinfo{author}{Sonnemans, J.}
\newblock \bibinfo{journal}{\bibinfo{title}{Learning by experience and learning
  by imitating successful others}}.
\newblock {\emph{\JournalTitle{Journal of Economic Behavior \& Organization}}}
  \textbf{\bibinfo{volume}{34}}, \bibinfo{pages}{559--575}
  (\bibinfo{year}{1998}).

\bibitem{jones2009development}
\bibinfo{author}{Jones, S.~S.}
\newblock \bibinfo{journal}{\bibinfo{title}{The development of imitation in
  infancy}}.
\newblock {\emph{\JournalTitle{Philosophical Transactions of the Royal Society
  B: Biological Sciences}}} \textbf{\bibinfo{volume}{364}},
  \bibinfo{pages}{2325--2335} (\bibinfo{year}{2009}).

\bibitem{pomerleaualvinn}
\bibinfo{author}{Pomerleau, D.~A.}
\newblock \bibinfo{journal}{\bibinfo{title}{Efficient training of artificial
  neural networks for autonomous navigation}}.
\newblock {\emph{\JournalTitle{Neural Computation}}}
  \textbf{\bibinfo{volume}{3}}, \bibinfo{pages}{88--97} (\bibinfo{year}{1991}).

\bibitem{abbeel2010autonomous}
\bibinfo{author}{Abbeel, P.}, \bibinfo{author}{Coates, A.} \&
  \bibinfo{author}{Ng, A.~Y.}
\newblock \bibinfo{journal}{\bibinfo{title}{Autonomous helicopter aerobatics
  through apprenticeship learning}}.
\newblock {\emph{\JournalTitle{The International Journal of Robotics
  Research}}} \textbf{\bibinfo{volume}{29}}, \bibinfo{pages}{1608--1639}
  (\bibinfo{year}{2010}).

\bibitem{uchibe2021forward}
\bibinfo{author}{Uchibe, E.} \& \bibinfo{author}{Doya, K.}
\newblock \bibinfo{journal}{\bibinfo{title}{Forward and inverse reinforcement
  learning sharing network weights and hyperparameters}}.
\newblock {\emph{\JournalTitle{Neural Networks}}}
  \textbf{\bibinfo{volume}{144}}, \bibinfo{pages}{138--153}
  (\bibinfo{year}{2021}).

\bibitem{scone2010trade}
\bibinfo{author}{Scone, S.} \& \bibinfo{author}{Phillips, I.}
\newblock \bibinfo{title}{Trade-off between exploration and reporting victim
  locations in usar}.
\newblock In \emph{\bibinfo{booktitle}{2010 IEEE International Symposium on" A
  World of Wireless, Mobile and Multimedia Networks"(WoWMoM)}},
  \bibinfo{pages}{1--6} (\bibinfo{organization}{IEEE}, \bibinfo{year}{2010}).

\bibitem{otte2013navigation}
\bibinfo{author}{Otte, M.}, \bibinfo{author}{Correll, N.} \&
  \bibinfo{author}{Frazzoli, E.}
\newblock \bibinfo{title}{Navigation with foraging}.
\newblock In \emph{\bibinfo{booktitle}{2013 IEEE/RSJ International Conference
  on Intelligent Robots and Systems}}, \bibinfo{pages}{3150--3157}
  (\bibinfo{organization}{IEEE}, \bibinfo{year}{2013}).

\bibitem{ccatal2021robot}
\bibinfo{author}{{\c{C}}atal, O.}, \bibinfo{author}{Verbelen, T.},
  \bibinfo{author}{Van~de Maele, T.}, \bibinfo{author}{Dhoedt, B.} \&
  \bibinfo{author}{Safron, A.}
\newblock \bibinfo{journal}{\bibinfo{title}{Robot navigation as hierarchical
  active inference}}.
\newblock {\emph{\JournalTitle{Neural Networks}}}
  \textbf{\bibinfo{volume}{142}}, \bibinfo{pages}{192--204}
  (\bibinfo{year}{2021}).

\bibitem{alexander2020egocentric}
\bibinfo{author}{Alexander, A.~S.} \emph{et~al.}
\newblock \bibinfo{journal}{\bibinfo{title}{Egocentric boundary vector tuning
  of the retrosplenial cortex}}.
\newblock {\emph{\JournalTitle{Science Advances}}}
  \textbf{\bibinfo{volume}{6}}, \bibinfo{pages}{eaaz2322}
  (\bibinfo{year}{2020}).

\bibitem{schaal1999imitation}
\bibinfo{author}{Schaal, S.}
\newblock \bibinfo{journal}{\bibinfo{title}{Is imitation learning the route to
  humanoid robots?}}
\newblock {\emph{\JournalTitle{Trends in Cognitive Sciences}}}
  \textbf{\bibinfo{volume}{3}}, \bibinfo{pages}{233--242}
  (\bibinfo{year}{1999}).

\bibitem{syed2010reduction}
\bibinfo{author}{Syed, U.} \& \bibinfo{author}{Schapire, R.~E.}
\newblock \bibinfo{journal}{\bibinfo{title}{A reduction from apprenticeship
  learning to classification}}.
\newblock {\emph{\JournalTitle{Advances in Neural Information Processing
  Systems}}} \textbf{\bibinfo{volume}{23}} (\bibinfo{year}{2010}).

\bibitem{ross2010efficient}
\bibinfo{author}{Ross, S.} \& \bibinfo{author}{Bagnell, D.}
\newblock \bibinfo{title}{Efficient reductions for imitation learning}.
\newblock In \emph{\bibinfo{booktitle}{Proceedings of the thirteenth
  International Conference on Artificial Intelligence and Statistics}},
  \bibinfo{pages}{661--668} (\bibinfo{organization}{JMLR Workshop and
  Conference Proceedings}, \bibinfo{year}{2010}).

\bibitem{osa2018algorithmic}
\bibinfo{author}{Osa, T.} \emph{et~al.}
\newblock \bibinfo{journal}{\bibinfo{title}{An algorithmic perspective on
  imitation learning}}.
\newblock {\emph{\JournalTitle{Foundations and Trends in Robotics}}}
  \textbf{\bibinfo{volume}{7}}, \bibinfo{pages}{1--179} (\bibinfo{year}{2018}).

\bibitem{sutton2018reinforcement}
\bibinfo{author}{Sutton, R.~S.} \& \bibinfo{author}{Barto, A.~G.}
\newblock \emph{\bibinfo{title}{Reinforcement learning: An introduction}}
  (\bibinfo{publisher}{MIT press}, \bibinfo{year}{2018}).

\bibitem{dulac2019challenges}
\bibinfo{author}{Dulac-Arnold, G.}, \bibinfo{author}{Mankowitz, D.} \&
  \bibinfo{author}{Hester, T.}
\newblock \bibinfo{journal}{\bibinfo{title}{Challenges of real-world
  reinforcement learning}}.
\newblock {\emph{\JournalTitle{arXiv preprint arXiv:1904.12901}}}
  (\bibinfo{year}{2019}).

\bibitem{ross2011reduction}
\bibinfo{author}{Ross, S.}, \bibinfo{author}{Gordon, G.} \&
  \bibinfo{author}{Bagnell, D.}
\newblock \bibinfo{title}{A reduction of imitation learning and structured
  prediction to no-regret online learning}.
\newblock In \emph{\bibinfo{booktitle}{Proceedings of the fourteenth
  International Conference on Artificial Intelligence and Statistics}},
  \bibinfo{pages}{627--635} (\bibinfo{organization}{JMLR Workshop and
  Conference Proceedings}, \bibinfo{year}{2011}).

\bibitem{ross2014reinforcement}
\bibinfo{author}{Ross, S.} \& \bibinfo{author}{Bagnell, J.~A.}
\newblock \bibinfo{journal}{\bibinfo{title}{Reinforcement and imitation
  learning via interactive no-regret learning}}.
\newblock {\emph{\JournalTitle{arXiv preprint arXiv:1406.5979}}}
  (\bibinfo{year}{2014}).

\bibitem{sun2018truncated}
\bibinfo{author}{Sun, W.}, \bibinfo{author}{Bagnell, J.~A.} \&
  \bibinfo{author}{Boots, B.}
\newblock \bibinfo{title}{Truncated horizon policy search: combining
  reinforcement learning and imitation learning}.
\newblock In \emph{\bibinfo{booktitle}{International Conference on Learning
  Representations}} (\bibinfo{year}{2018}).

\bibitem{cheng2019fast}
\bibinfo{author}{Cheng, C.}, \bibinfo{author}{Yan, X.},
  \bibinfo{author}{Wagener, N.} \& \bibinfo{author}{Boots, B.}
\newblock \bibinfo{title}{Fast policy learning through imitation and
  reinforcement}.
\newblock In \emph{\bibinfo{booktitle}{Uncertainty in Artificial Intelligence}}
  (\bibinfo{year}{2019}).

\bibitem{subramanian2016exploration}
\bibinfo{author}{Subramanian, K.}, \bibinfo{author}{Isbell~Jr, C.~L.} \&
  \bibinfo{author}{Thomaz, A.~L.}
\newblock \bibinfo{title}{Exploration from demonstration for interactive
  reinforcement learning}.
\newblock In \emph{\bibinfo{booktitle}{Proceedings of the 2016 International
  Conference on Autonomous Agents \& Multiagent Systems}},
  \bibinfo{pages}{447--456} (\bibinfo{year}{2016}).

\bibitem{vecerik2017leveraging}
\bibinfo{author}{Vecerik, M.} \emph{et~al.}
\newblock \bibinfo{journal}{\bibinfo{title}{Leveraging demonstrations for deep
  reinforcement learning on robotics problems with sparse rewards}}.
\newblock {\emph{\JournalTitle{arXiv preprint arXiv:1707.08817}}}
  (\bibinfo{year}{2017}).

\bibitem{nair2018overcoming}
\bibinfo{author}{Nair, A.}, \bibinfo{author}{McGrew, B.},
  \bibinfo{author}{Andrychowicz, M.}, \bibinfo{author}{Zaremba, W.} \&
  \bibinfo{author}{Abbeel, P.}
\newblock \bibinfo{title}{Overcoming exploration in reinforcement learning with
  demonstrations}.
\newblock In \emph{\bibinfo{booktitle}{2018 IEEE International Conference on
  Robotics and Automation (ICRA)}}, \bibinfo{pages}{6292--6299}
  (\bibinfo{organization}{IEEE}, \bibinfo{year}{2018}).

\bibitem{libardi2021guided}
\bibinfo{author}{Libardi, G.}, \bibinfo{author}{De~Fabritiis, G.} \&
  \bibinfo{author}{Dittert, S.}
\newblock \bibinfo{title}{Guided exploration with proximal policy optimization
  using a single demonstration}.
\newblock In \emph{\bibinfo{booktitle}{International Conference on Machine
  Learning}}, \bibinfo{pages}{6611--6620} (\bibinfo{organization}{PMLR},
  \bibinfo{year}{2021}).

\bibitem{Uchendu2021DemonstrationGuidedQ}
\bibinfo{author}{Uchendu, I.} \emph{et~al.}
\newblock \bibinfo{journal}{\bibinfo{title}{Demonstration-guided q-learning}}.
\newblock {\emph{\JournalTitle{NIPS Workshop on Robot Learning: Self-Supervised
  and Lifelong Learning}}}  (\bibinfo{year}{2021}).

\bibitem{abbeel2004apprenticeship}
\bibinfo{author}{Abbeel, P.} \& \bibinfo{author}{Ng, A.~Y.}
\newblock \bibinfo{title}{Apprenticeship learning via inverse reinforcement
  learning}.
\newblock In \emph{\bibinfo{booktitle}{Proceedings of the twenty-first
  International Conference on Machine Learning}}, \bibinfo{pages}{1}
  (\bibinfo{year}{2004}).

\bibitem{ratliff2006maximum}
\bibinfo{author}{Ratliff, N.~D.}, \bibinfo{author}{Bagnell, J.~A.} \&
  \bibinfo{author}{Zinkevich, M.~A.}
\newblock \bibinfo{title}{Maximum margin planning}.
\newblock In \emph{\bibinfo{booktitle}{Proceedings of the twenty-third
  International Conference on Machine learning}}, \bibinfo{pages}{729--736}
  (\bibinfo{year}{2006}).

\bibitem{ziebart2008maximum}
\bibinfo{author}{Ziebart, B.~D.}, \bibinfo{author}{Maas, A.~L.},
  \bibinfo{author}{Bagnell, J.~A.}, \bibinfo{author}{Dey, A.~K.} \emph{et~al.}
\newblock \bibinfo{title}{Maximum entropy inverse reinforcement learning.}
\newblock In \emph{\bibinfo{booktitle}{Proceedings of the AAAI Conference on
  Artificial Intelligence}}, vol.~\bibinfo{volume}{8},
  \bibinfo{pages}{1433--1438} (\bibinfo{organization}{Chicago, IL, USA},
  \bibinfo{year}{2008}).

\bibitem{finn2016guided}
\bibinfo{author}{Finn, C.}, \bibinfo{author}{Levine, S.} \&
  \bibinfo{author}{Abbeel, P.}
\newblock \bibinfo{title}{Guided cost learning: Deep inverse optimal control
  via policy optimization}.
\newblock In \emph{\bibinfo{booktitle}{International Conference on Machine
  Learning}}, \bibinfo{pages}{49--58} (\bibinfo{organization}{PMLR},
  \bibinfo{year}{2016}).

\bibitem{ho2016generative}
\bibinfo{author}{Ho, J.} \& \bibinfo{author}{Ermon, S.}
\newblock \bibinfo{journal}{\bibinfo{title}{Generative adversarial imitation
  learning}}.
\newblock {\emph{\JournalTitle{Advances in Neural Information Processing
  Systems}}} \textbf{\bibinfo{volume}{29}} (\bibinfo{year}{2016}).

\bibitem{ghasemipour2020divergence}
\bibinfo{author}{Ghasemipour, S. K.~S.}, \bibinfo{author}{Zemel, R.} \&
  \bibinfo{author}{Gu, S.}
\newblock \bibinfo{title}{A divergence minimization perspective on imitation
  learning methods}.
\newblock In \emph{\bibinfo{booktitle}{Conference on Robot Learning}},
  \bibinfo{pages}{1259--1277} (\bibinfo{organization}{PMLR},
  \bibinfo{year}{2020}).

\bibitem{goodfellow2020generative}
\bibinfo{author}{Goodfellow, I.} \emph{et~al.}
\newblock \bibinfo{journal}{\bibinfo{title}{Generative adversarial networks}}.
\newblock {\emph{\JournalTitle{Communications of the ACM}}}
  \textbf{\bibinfo{volume}{63}}, \bibinfo{pages}{139--144}
  (\bibinfo{year}{2020}).

\bibitem{kang2018policy}
\bibinfo{author}{Kang, B.}, \bibinfo{author}{Jie, Z.} \& \bibinfo{author}{Feng,
  J.}
\newblock \bibinfo{title}{Policy optimization with demonstrations}.
\newblock In \emph{\bibinfo{booktitle}{International Conference on Machine
  Learning}}, \bibinfo{pages}{2469--2478} (\bibinfo{organization}{PMLR},
  \bibinfo{year}{2018}).

\bibitem{hester2018deep}
\bibinfo{author}{Hester, T.} \emph{et~al.}
\newblock \bibinfo{title}{Deep q-learning from demonstrations}.
\newblock In \emph{\bibinfo{booktitle}{Proceedings of the AAAI Conference on
  Artificial Intelligence}}, vol.~\bibinfo{volume}{32} (\bibinfo{year}{2018}).

\bibitem{mnih2013playing}
\bibinfo{author}{Mnih, V.} \emph{et~al.}
\newblock \bibinfo{journal}{\bibinfo{title}{Playing atari with deep
  reinforcement learning}}.
\newblock {\emph{\JournalTitle{arXiv preprint arXiv:1312.5602}}}
  (\bibinfo{year}{2013}).

\bibitem{levine2020offline}
\bibinfo{author}{Levine, S.}, \bibinfo{author}{Kumar, A.},
  \bibinfo{author}{Tucker, G.} \& \bibinfo{author}{Fu, J.}
\newblock \bibinfo{journal}{\bibinfo{title}{Offline reinforcement learning:
  Tutorial, review, and perspectives on open problems}}.
\newblock {\emph{\JournalTitle{arXiv preprint arXiv:2005.01643}}}
  (\bibinfo{year}{2020}).

\bibitem{Moore2021virtual}
\bibinfo{author}{Moore, K.}, \bibinfo{author}{Yi, C.}, \bibinfo{author}{Dunne,
  M.}, \bibinfo{author}{Stern, C.} \& \bibinfo{author}{McGuire, J.}
\newblock \bibinfo{journal}{\bibinfo{title}{Virtual human foraging behavior
  follows predictions for heavy-tailed search.}}
\newblock {\emph{\JournalTitle{In Society for Neuroscience}}}
  \textbf{\bibinfo{volume}{Online}} (\bibinfo{year}{2021}).

\bibitem{feigenbaum2004allocentric}
\bibinfo{author}{Feigenbaum, J.~D.} \& \bibinfo{author}{Morris, R.~G.}
\newblock \bibinfo{journal}{\bibinfo{title}{Allocentric versus egocentric
  spatial memory after unilateral temporal lobectomy in humans.}}
\newblock {\emph{\JournalTitle{Neuropsychology}}}
  \textbf{\bibinfo{volume}{18}}, \bibinfo{pages}{462} (\bibinfo{year}{2004}).

\bibitem{schulman2015trust}
\bibinfo{author}{Schulman, J.}, \bibinfo{author}{Levine, S.},
  \bibinfo{author}{Abbeel, P.}, \bibinfo{author}{Jordan, M.} \&
  \bibinfo{author}{Moritz, P.}
\newblock \bibinfo{title}{Trust region policy optimization}.
\newblock In \emph{\bibinfo{booktitle}{International Conference on Machine
  Learning}}, \bibinfo{pages}{1889--1897} (\bibinfo{organization}{PMLR},
  \bibinfo{year}{2015}).

\bibitem{queeney2021uncertainty}
\bibinfo{author}{Queeney, J.}, \bibinfo{author}{Paschalidis, I.~C.} \&
  \bibinfo{author}{Cassandras, C.~G.}
\newblock \bibinfo{title}{Uncertainty-aware policy optimization: A robust,
  adaptive trust region approach}.
\newblock In \emph{\bibinfo{booktitle}{Proceedings of the AAAI Conference on
  Artificial Intelligence}}, vol.~\bibinfo{volume}{35},
  \bibinfo{pages}{9377--9385} (\bibinfo{year}{2021}).

\bibitem{haarnoja2018soft}
\bibinfo{author}{Haarnoja, T.}, \bibinfo{author}{Zhou, A.},
  \bibinfo{author}{Abbeel, P.} \& \bibinfo{author}{Levine, S.}
\newblock \bibinfo{title}{Soft actor-critic: Off-policy maximum entropy deep
  reinforcement learning with a stochastic actor}.
\newblock In \emph{\bibinfo{booktitle}{International Conference on Machine
  Learning}}, \bibinfo{pages}{1861--1870} (\bibinfo{organization}{PMLR},
  \bibinfo{year}{2018}).

\bibitem{fujimoto2018addressing}
\bibinfo{author}{Fujimoto, S.}, \bibinfo{author}{Hoof, H.} \&
  \bibinfo{author}{Meger, D.}
\newblock \bibinfo{title}{Addressing function approximation error in
  actor-critic methods}.
\newblock In \emph{\bibinfo{booktitle}{International Conference on Machine
  Learning}}, \bibinfo{pages}{1587--1596} (\bibinfo{organization}{PMLR},
  \bibinfo{year}{2018}).

\bibitem{andrychowicz2020matters}
\bibinfo{author}{Andrychowicz, M.} \emph{et~al.}
\newblock \bibinfo{journal}{\bibinfo{title}{What matters in on-policy
  reinforcement learning? a large-scale empirical study}}.
\newblock {\emph{\JournalTitle{arXiv preprint arXiv:2006.05990}}}
  (\bibinfo{year}{2020}).

\bibitem{engstrom2020implementation}
\bibinfo{author}{Engstrom, L.} \emph{et~al.}
\newblock \bibinfo{title}{Implementation matters in deep policy gradients: A
  case study on ppo and trpo}.
\newblock In \emph{\bibinfo{booktitle}{International Conference on Learning
  Representations}} (\bibinfo{year}{2020}).

\bibitem{sutton1999policy}
\bibinfo{author}{Sutton, R.~S.}, \bibinfo{author}{McAllester, D.~A.},
  \bibinfo{author}{Singh, S.~P.}, \bibinfo{author}{Mansour, Y.} \emph{et~al.}
\newblock \bibinfo{title}{Policy gradient methods for reinforcement learning
  with function approximation.}
\newblock In \emph{\bibinfo{booktitle}{NIPs}}, vol.~\bibinfo{volume}{99},
  \bibinfo{pages}{1057--1063} (\bibinfo{organization}{Citeseer},
  \bibinfo{year}{1999}).

\bibitem{queeney2021generalized}
\bibinfo{author}{Queeney, J.}, \bibinfo{author}{Paschalidis, I.} \&
  \bibinfo{author}{Cassandras, C.}
\newblock \bibinfo{journal}{\bibinfo{title}{Generalized proximal policy
  optimization with sample reuse}}.
\newblock {\emph{\JournalTitle{Advances in Neural Information Processing
  Systems}}} \textbf{\bibinfo{volume}{34}} (\bibinfo{year}{2021}).

\bibitem{kakade2002approximately}
\bibinfo{author}{Kakade, S.} \& \bibinfo{author}{Langford, J.}
\newblock \bibinfo{title}{Approximately optimal approximate reinforcement
  learning}.
\newblock In \emph{\bibinfo{booktitle}{In Proc. 19th International Conference
  on Machine Learning}} (\bibinfo{organization}{Citeseer},
  \bibinfo{year}{2002}).

\end{thebibliography}

\clearpage
\begin{appendix}

\section*{Appendix}

\subsection*{On-policy vs.\ off-policy RL algorithms}

The Dichotomy between on-policy and off-policy reinforcement learning (RL) algorithms started since the earliest days of RL and specifically with temporal-difference (TD) learning.\cite{sutton2018reinforcement}.

Recall that the goal of RL is to find $\bm{\theta}$ such that the expected total discounted reward $J(\bm{\theta})=\mathbb{E}_{\tau}[\sum_{t=0}^{\infty}\gamma^t r(s_t,a_t)]$ is maximized and where $\tau = (s_0,a_0,s_1,a_1,\dots)$ is sampled according to $s_0 \sim D$, $a_t\sim\pi_{\bm{\theta}}(\cdot|s_t)$ and $s_{t+1}\sim P(\cdot|s_t,a_t)$. The mechanism used in all the mentioned algorithms consists of alternating policy evaluation with policy improvement and doing it for several iterations until convergence. Specifically, the policy evaluation step consists of using the agent's policy $\pi_{\bm{\theta}}$ to generate trajectories $\tau = (s_0,a_0,s_1,a_1,\dots)$ to estimate the state-action value function denoted by $Q_{\bm{\theta}}(s,a) = \mathbb{E}_{\tau}[\sum_{t=0}^{\infty}\gamma^t r(s_t,a_t)|S_0=s, A_0=a]$, and/or the state value function $V_{\bm{\theta}}(s)=\mathbb{E}_{a \sim \pi_{\bm{\theta}}(\cdot|s_t)}[Q_{\bm{\theta}}(s,a)]$. Then, the policy improvement step, updates the current policy $\pi_{\bm{\theta}}$ via a policy gradient in a direction that maximizes $J(\bm{\theta})$. \cite{sutton1999policy}

On-policy algorithms rely only on trajectories $\tau$ generated by the interaction of $\pi_{\bm{\theta}}$ with the environment at the current iteration. Whereas, off-policy algorithms retrieve trajectories generated also at previous iterations and use this past experience in the policy update. This has indeed a different connotation in practice.\cite{queeney2021generalized} On-policy methods deliver stable performance throughout learning due to their connection to theoretical policy improvement guarantees.\cite{kakade2002approximately} However, theoretically supported stability yields high sample complexity due to the conspicuous amount of trajectories needed at each iteration. On the other hand, off-policy algorithms address the sample complexity issue by storing data in a replay buffer which are then reused for multiple policy updates. Off-policy methods have proved to be more sample efficient in practice but also more unstable due to the bias introduced in estimating the state-action value function $Q_{\bm{\theta}}(s,a)$. A recent work has sought to combine on-policy algorithms with off-policy ideas.\cite{queeney2021generalized}

\begin{figure}[ht]
    \centering
    \includegraphics[width=11cm,height=4cm]{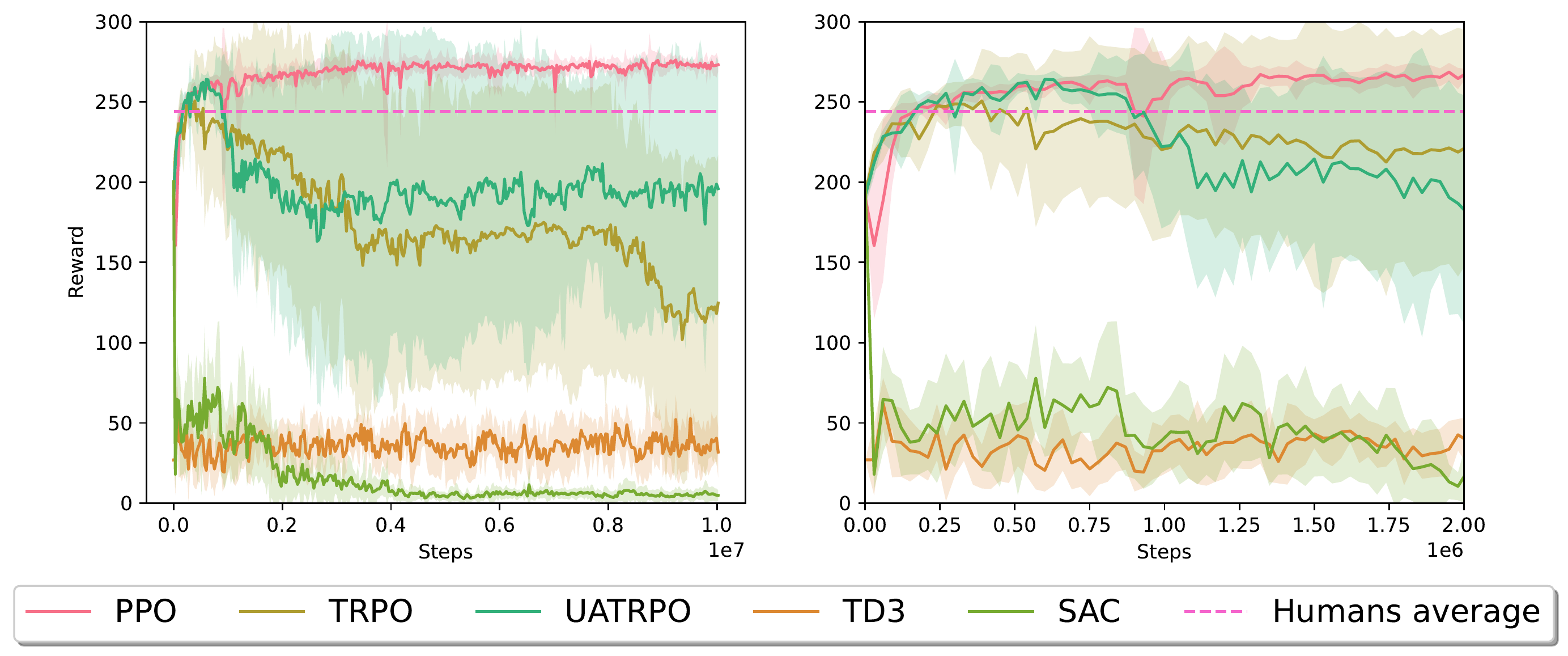}
    \caption{The results for the RL algorithms initialized with the same imitation learning policy.}
    \label{fig_app:RL_algortihms_comparison}
\end{figure}

In the specific case of this paper, we observe that the instability due to off-policy updates can undermine the imitation learning (IL) initialization as illustrated in Fig.~\ref{fig_app:RL_algortihms_comparison} (see SAC and TD3). At the same time, the IL initialization remarkably helps on-policy methods to accomplish fast and efficient convergence. Thus, IL+PPO results in a more reliable and efficient learning method. 

\subsection*{Neural Network design}

We use the same $\pi_{\bm{\theta}}$ architecture for all the algorithms. On-policy algorithms rely on Generalized Advantage Estimation to compute their critic, this method trade-offs between high variance Monte-Carlo estimation and biased Bootstrap and relies only on the value function network $V_{\bm{\theta}_v}$. On the other hand, off-policy algorithms use TD learning which relies on a state-action value function $Q_{\bm{\theta}_q}$. All the architectures are summarized in Table~\ref{tab:NN}.

\begin{table}[h!]
\centering
\caption{Neural Networks architectures.}
\label{tab:NN}
\small
\begin{tabular}{c c c c c c}\toprule
\multicolumn{6}{l}{$\pi_{\bm{\theta}}$ network architecture}\\
\midrule
\multicolumn{2}{c}{Layer type} & \multicolumn{2}{c}{Dimension} & \multicolumn{2}{c}{Activation Function} \\
\cmidrule(lr){1-2} \cmidrule(lr){3-4} \cmidrule(lr){5-6}
\multicolumn{2}{c}{Input} & \multicolumn{2}{c}{observation size} & \multicolumn{2}{c}{Linear} \\
\multicolumn{2}{c}{Hidden fully connected} & \multicolumn{2}{c}{128} & \multicolumn{2}{c}{Relu} \\
\multicolumn{2}{c}{Output} & \multicolumn{2}{c}{action space cardinality} & \multicolumn{2}{c}{Softmax} \\
& & & & & \\
\midrule
\multicolumn{6}{l}{$V_{\bm{\theta}_v}$ network architecture.}\\
\midrule
\multicolumn{2}{c}{Layer type} & \multicolumn{2}{c}{Dimension} & \multicolumn{2}{c}{Activation Function} \\
\cmidrule(lr){1-2} \cmidrule(lr){3-4} \cmidrule(lr){5-6}
\multicolumn{2}{c}{Input} & \multicolumn{2}{c}{observation size} & \multicolumn{2}{c}{Linear} \\
\multicolumn{2}{c}{Hidden fully connected} & \multicolumn{2}{c}{256} & \multicolumn{2}{c}{Relu} \\
\multicolumn{2}{c}{Output} & \multicolumn{2}{c}{1} & \multicolumn{2}{c}{Linear} \\
& & & & & \\
\midrule
\multicolumn{6}{l}{$Q_{\bm{\theta}_q}$ network architecture.}\\
\midrule
\multicolumn{2}{c}{Layer type} & \multicolumn{2}{c}{Dimension} & \multicolumn{2}{c}{Activation Function} \\
\cmidrule(lr){1-2} \cmidrule(lr){3-4} \cmidrule(lr){5-6}
\multicolumn{2}{c}{Input} & \multicolumn{2}{c}{observation size} & \multicolumn{2}{c}{Linear} \\
\multicolumn{2}{c}{Hidden fully connected} & \multicolumn{2}{c}{256} & \multicolumn{2}{c}{Relu} \\
\multicolumn{2}{c}{Output} & \multicolumn{2}{c}{action space cardinality} & \multicolumn{2}{c}{Linear} \\
\bottomrule
\end{tabular}
\end{table}

\subsection*{Hyperparameter values, human trajectories and experiments}

\begin{table}[h!]
\centering
\caption{Hyperparameter values for experimental setup.}
\label{tab:Hyper_1}
\small
\begin{tabular}{c c c c}\toprule
\multicolumn{2}{l}{Evaluation} & \multicolumn{2}{c}{Default} \\
\cmidrule(lr){1-2} \cmidrule(lr){3-4}
\multicolumn{2}{l}{Episodes per evaluations} & \multicolumn{2}{c}{10} \\
\multicolumn{2}{l}{Steps per evaluation episode} & \multicolumn{2}{c}{3464} \\
& & & \\
\multicolumn{2}{l}{General IL} & &\\
\cmidrule(lr){1-2}
\multicolumn{2}{l}{Optimization algorithm} & \multicolumn{2}{c}{Minibatch gradient ascent} \\
\multicolumn{2}{l}{Size minibatches} & \multicolumn{2}{c}{32} \\
\multicolumn{2}{l}{Epochs per update} & \multicolumn{2}{c}{1} \\
\multicolumn{2}{l}{Number of updates} & \multicolumn{2}{c}{11}\\
\multicolumn{2}{l}{Optimizer} & \multicolumn{2}{c}{Adam}\\
\multicolumn{2}{l}{Learning rate} & \multicolumn{2}{c}{0.001}\\
& & & \\
\multicolumn{2}{l}{General RL on-policy} & & \\
\cmidrule(lr){1-2}
\multicolumn{2}{l}{Max number of steps} & \multicolumn{2}{c}{$10.02 \times 10^6$} \\
\multicolumn{2}{l}{Steps between evaluations} & \multicolumn{2}{c}{$30000$}\\
\multicolumn{2}{l}{GAE $\gamma$} & \multicolumn{2}{c}{0.99} \\
\multicolumn{2}{l}{GAE $\lambda$} & \multicolumn{2}{c}{0.99} \\
& & & \\
\multicolumn{2}{l}{PPO} & & \\
\cmidrule(lr){1-2}
\multicolumn{2}{l}{Clipping parameter $\epsilon$} & \multicolumn{2}{c}{0.2} \\
\multicolumn{2}{l}{Entropy} & \multicolumn{2}{c}{True} \\
\multicolumn{2}{l}{Entropy weight $c_2$} & \multicolumn{2}{c}{$10^{-2}$}\\
\multicolumn{2}{l}{Optimization algorithm} & \multicolumn{2}{c}{Minibatch gradient ascent} \\
\multicolumn{2}{l}{Batch size} & \multicolumn{2}{c}{$30000$}\\
\multicolumn{2}{l}{Size minibatches} & \multicolumn{2}{c}{64} \\
\multicolumn{2}{l}{Epochs per update} & \multicolumn{2}{c}{10} \\
\multicolumn{2}{l}{Optimizer} & \multicolumn{2}{c}{Adam}\\
\multicolumn{2}{l}{Learning rate} & \multicolumn{2}{c}{$3 \times 10^{-4}$} \\
& & & \\
\multicolumn{2}{l}{TRPO} & & \\
\cmidrule(lr){1-2}
\multicolumn{2}{l}{Trust region parameter $\epsilon$} & \multicolumn{2}{c}{0.05} \\
\multicolumn{2}{l}{Entropy} & \multicolumn{2}{c}{False} \\
\multicolumn{2}{l}{Optimization algorithm} & \multicolumn{2}{c}{Conjugate gradient ascent with line search} \\
\multicolumn{2}{l}{Conjugate gradient damping} & \multicolumn{2}{c}{$0.1$} \\
\multicolumn{2}{l}{Conjugate iterations per update} & \multicolumn{2}{c}{$20$} \\
& & & \\
\multicolumn{2}{l}{UATRPO} & & \\
\cmidrule(lr){1-2}
\multicolumn{2}{l}{Trust region parameter $\epsilon$} & \multicolumn{2}{c}{0.03} \\
\multicolumn{2}{l}{Entropy} & \multicolumn{2}{c}{False} \\
\multicolumn{2}{l}{Optimization algorithm} & \multicolumn{2}{c}{Conjugate gradient ascent with line search} \\
\multicolumn{2}{l}{Conjugate gradient damping} & \multicolumn{2}{c}{$0.1$} \\
\multicolumn{2}{l}{Conjugate gradient iterations per update} & \multicolumn{2}{c}{$20$} \\
\multicolumn{2}{l}{Trade-off Parameter (c)} & \multicolumn{2}{c}{$6 \times 10^{-4}$} \\
\multicolumn{2}{l}{Confidence parameter ($\alpha$)} & \multicolumn{2}{c}{$0.05$} \\
\multicolumn{2}{l}{Number of random projections ($m$)} & \multicolumn{2}{c}{$200$} \\
\multicolumn{2}{l}{EMA weight parameter ($\beta$)} & \multicolumn{2}{c}{$0.9$} \\
\bottomrule
\end{tabular}
\end{table}

\begin{table}[h!]
\centering
\caption{Hyperparameter values for experimental setup (Cont.).}
\label{tab:Hyper_2}
\small
\begin{tabular}{c c c c}\toprule
\multicolumn{2}{l}{General RL off-policy} & \multicolumn{2}{c}{Default} \\
\cmidrule(lr){1-2} \cmidrule(lr){3-4}
\multicolumn{2}{l}{Max number of steps} & \multicolumn{2}{c}{$10.02 \times 10^6$} \\
\multicolumn{2}{l}{Steps between evaluations} & \multicolumn{2}{c}{$30000$}\\
\multicolumn{2}{l}{Buffer size} & \multicolumn{2}{c}{$10^{6}$} \\
\multicolumn{2}{l}{Discount $\gamma$} & \multicolumn{2}{c}{0.99} \\
\multicolumn{2}{l}{Optimization algorithm} & \multicolumn{2}{c}{Batch gradient ascent} \\
\multicolumn{2}{l}{Batch size} & \multicolumn{2}{c}{$256$}\\
\multicolumn{2}{l}{Optimizer} & \multicolumn{2}{c}{Adam}\\
\multicolumn{2}{l}{Learning rate} & \multicolumn{2}{c}{$3 \times 10^{-4}$} \\
& & & \\
\multicolumn{2}{l}{SAC} & & \\
\cmidrule(lr){1-2}
\multicolumn{2}{l}{Entropy temperature init $\alpha$} & \multicolumn{2}{c}{$0.2$}\\
\multicolumn{2}{l}{Target critic update step ($\tau$)} & \multicolumn{2}{c}{$0.005$} \\
& & & \\
\multicolumn{2}{l}{TD3} & & \\
\cmidrule(lr){1-2}
\multicolumn{2}{l}{Exploration noise std} & \multicolumn{2}{c}{$0.1$}\\
\multicolumn{2}{l}{Policy Noise std} & \multicolumn{2}{c}{$0.2$} \\
\multicolumn{2}{l}{Noise clip} & \multicolumn{2}{c}{$0.5$} \\
\multicolumn{2}{l}{Update steps delay} & \multicolumn{2}{c}{$2$} \\
\multicolumn{2}{l}{Target critic and policy  update step ($\tau$)} & \multicolumn{2}{c}{$0.005$}\\
\bottomrule
\end{tabular}
\end{table}

\begin{figure}
    \centering
    \includegraphics[width=0.7\linewidth,height = 18cm]{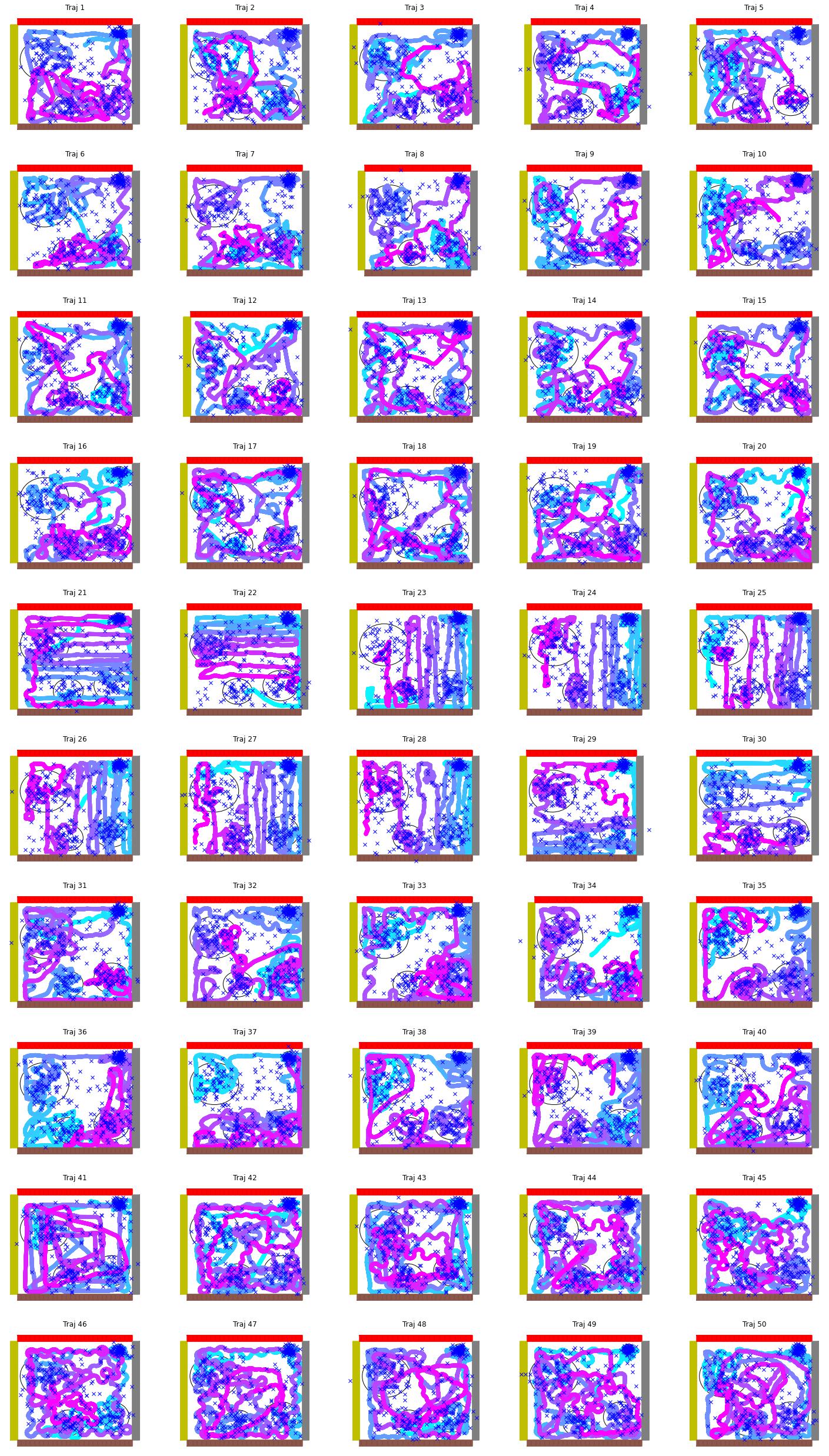}
    \caption{The complete set of $50$ 8-minutes human trajectories collected during the $2^{\text{nd}}$ day of tests.}
    \label{fig:human_trajs}
\end{figure}
\newpage

\begin{figure}
    \centering
    \includegraphics[width=0.7\linewidth,height = 18cm]{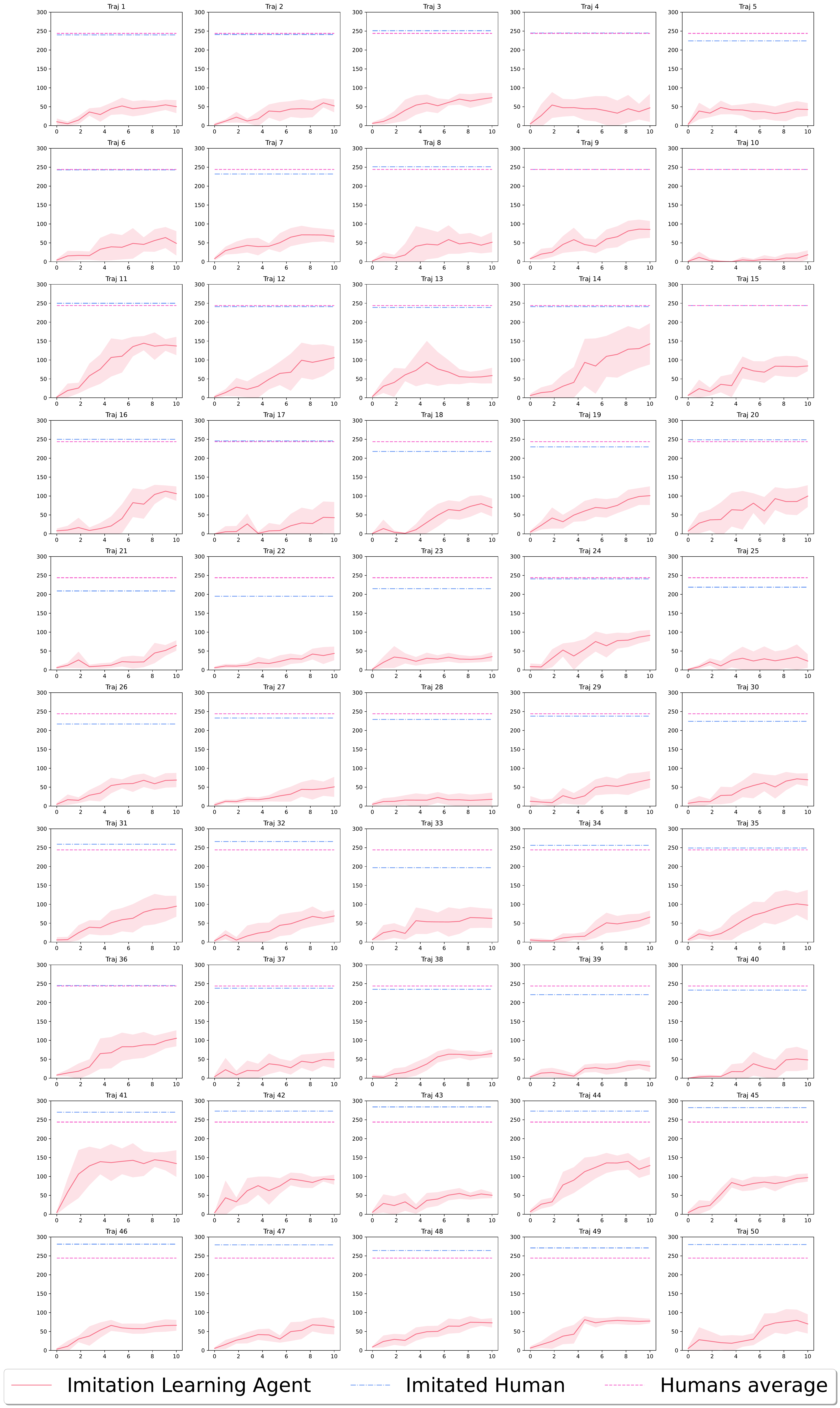}
    \caption{Results of the IL step for all the $50$ human trajectories in Fig.~\ref{fig:human_trajs}.}
    \label{fig:IL}
\end{figure}

\begin{figure}
    \centering
    \includegraphics[width=0.7\linewidth,height = 18cm]{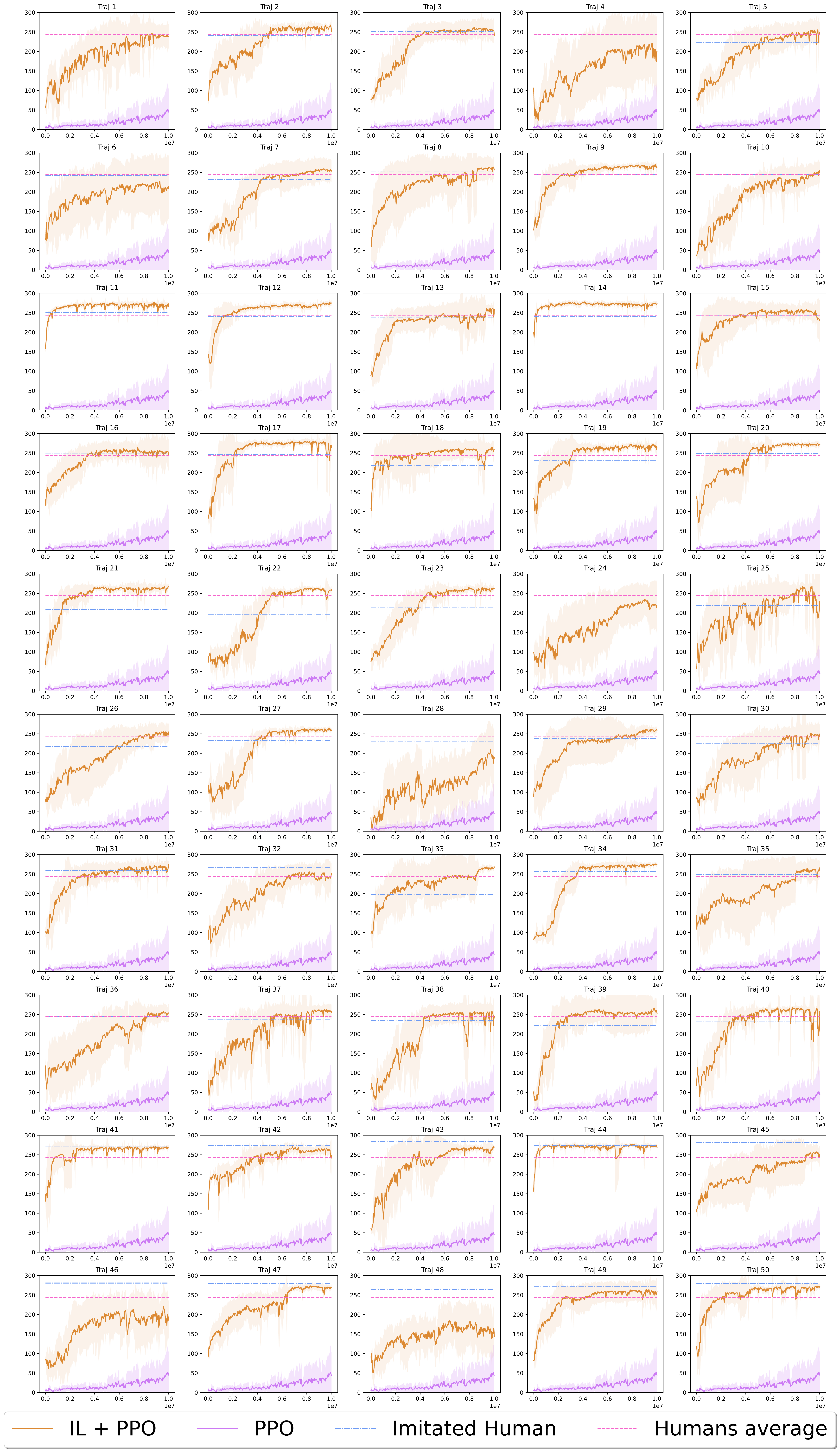}
    \caption{Results of the PPO step with the $50$ IL policies in Fig.~\ref{fig:IL}.}
    \label{fig:IL+RL}
\end{figure}

\begin{figure}
    \centering
    \includegraphics[width=0.7\linewidth,height = 18cm]{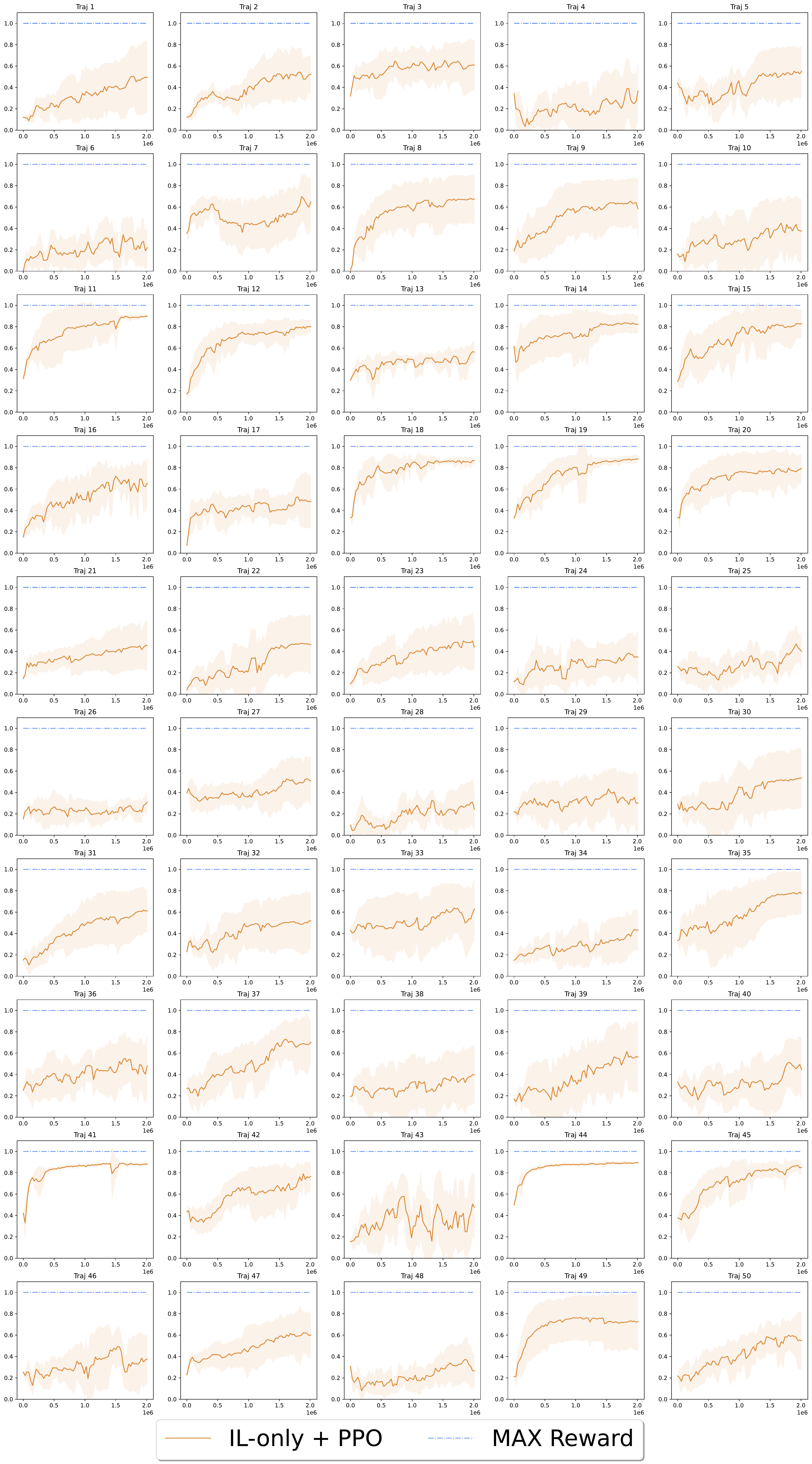}
    \caption{Results of learning a new rewards distribution in Fig~\ref{fig:top_view_ADV} with the IL-only initialization for all the $50$ IL policies in Fig.~\ref{fig:IL}.}
    \label{fig:IL_init_ADV}
\end{figure}

\begin{figure}
    \centering
    \includegraphics[width=0.7\linewidth,height = 18cm]{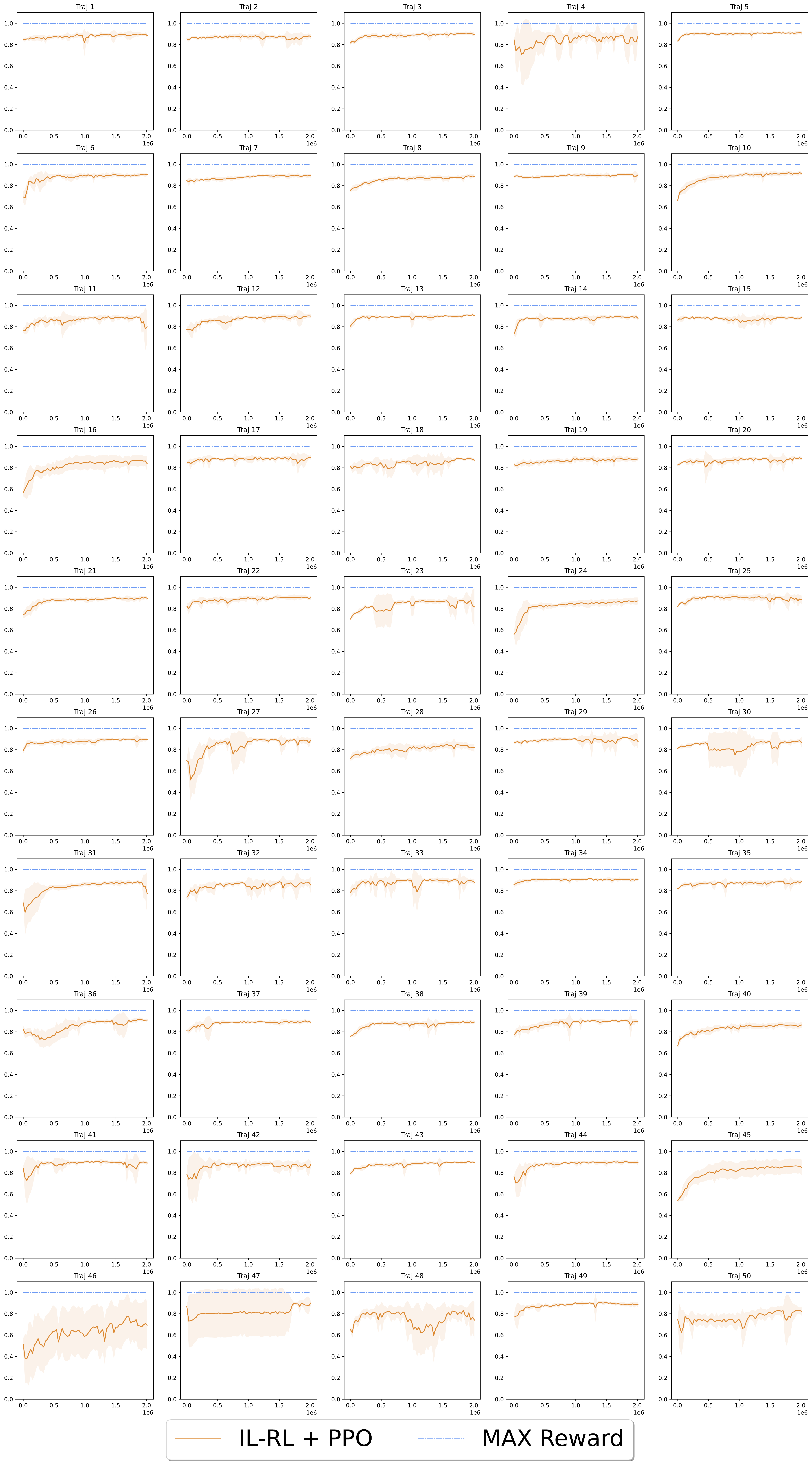}
    \caption{Results of learning a new rewards distribution in Fig~\ref{fig:top_view_ADV} with IL+PPO initialization for all the $50$ policies in Fig.~\ref{fig:IL+RL}.}
    \label{fig:IL+RL_init_ADV}
\end{figure}

\begin{figure}
    \centering
    \includegraphics[width=0.7\linewidth,height = 18cm]{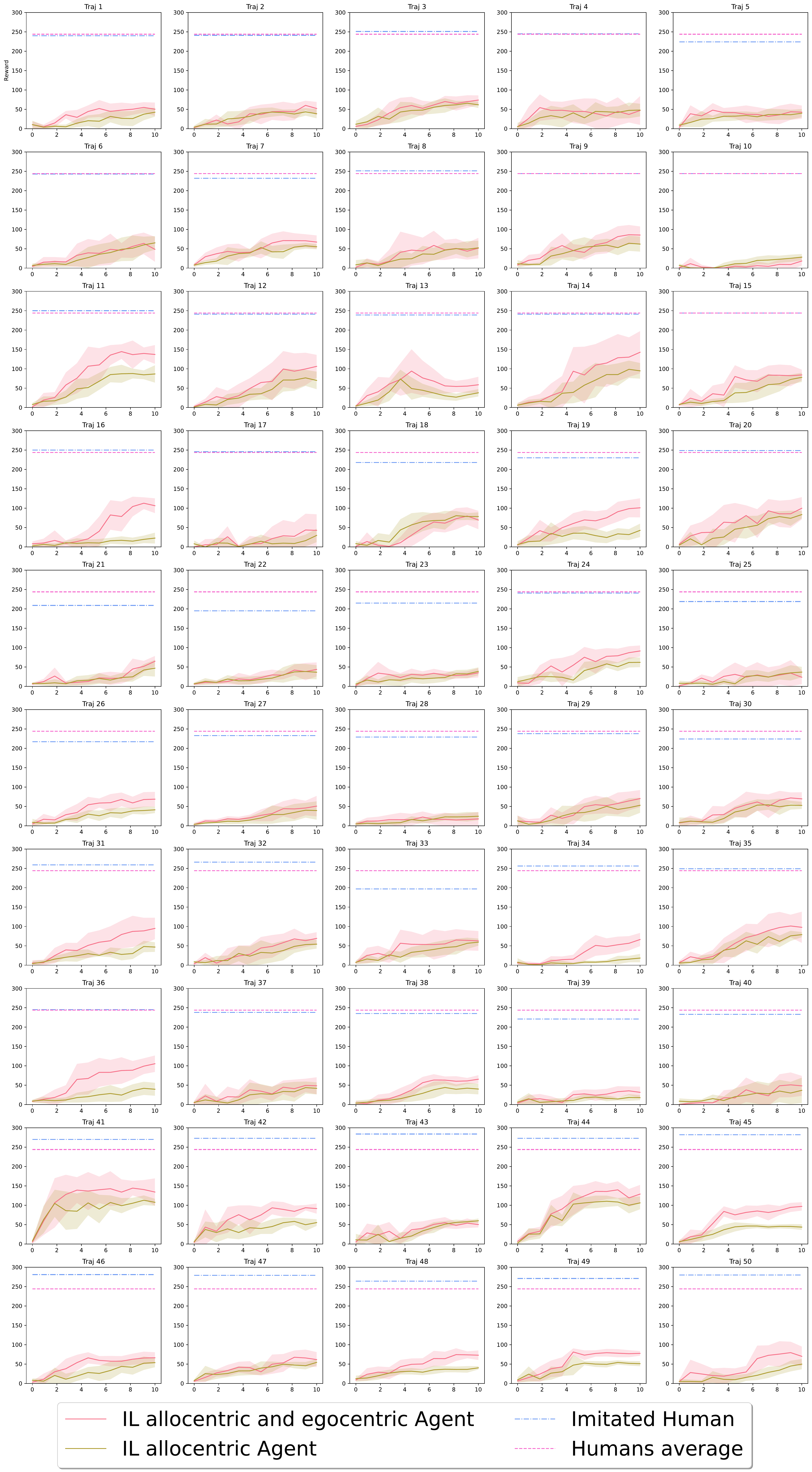}
    \caption{Comparison between IL with full state and allocentric only state for all the $50$ human trajectories.}
    \label{fig:IL_allocentric}
\end{figure}

\begin{figure}
    \centering
    \includegraphics[width=0.7\linewidth,height = 18cm]{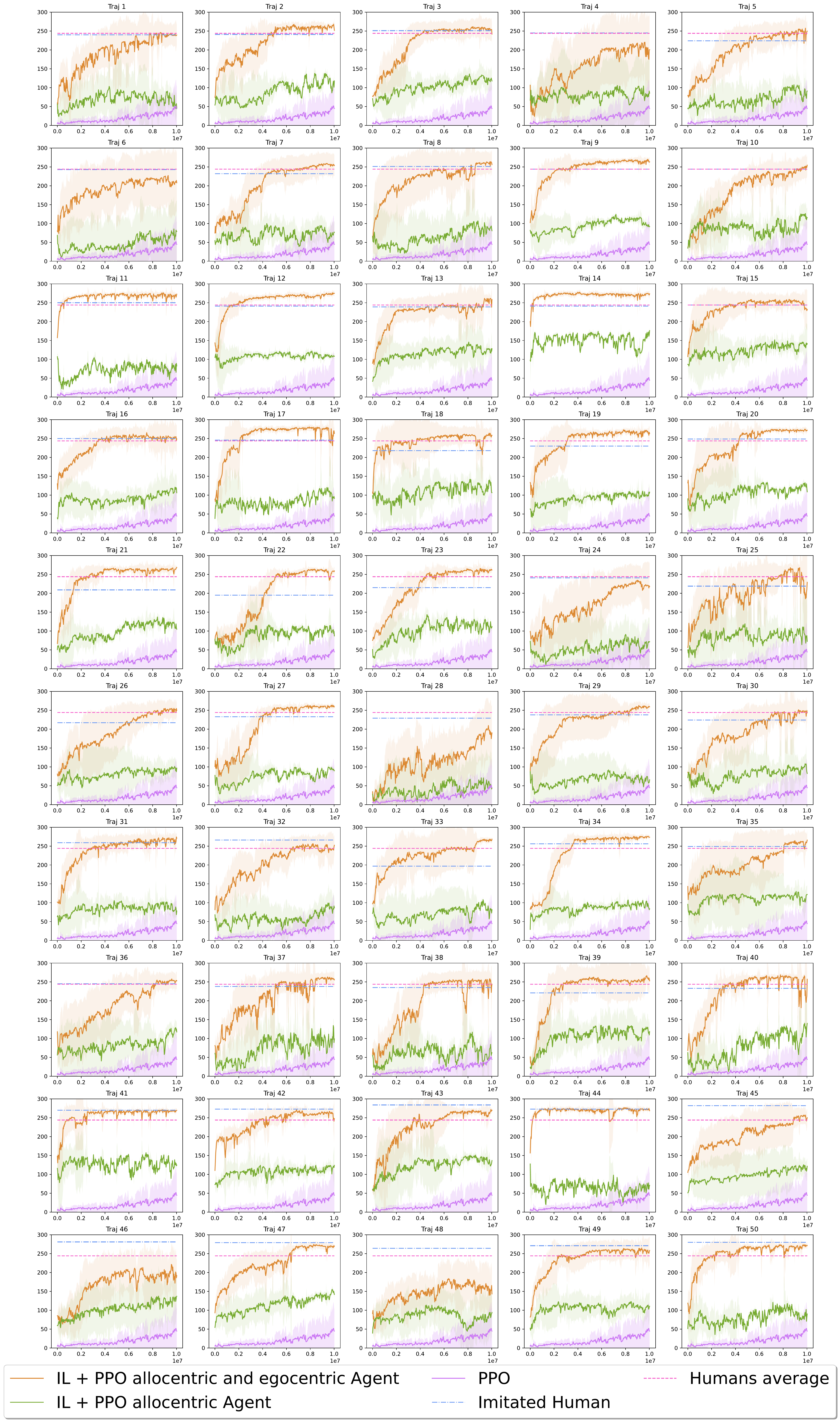}
    \caption{Comparison between IL+PPO with full state and allocentric only state for all the $50$ human trajectories.}
    \label{fig:IL+RL_allocentric}
\end{figure}

\begin{figure}
    \centering
    \includegraphics[width=0.8\linewidth,height = 20cm]{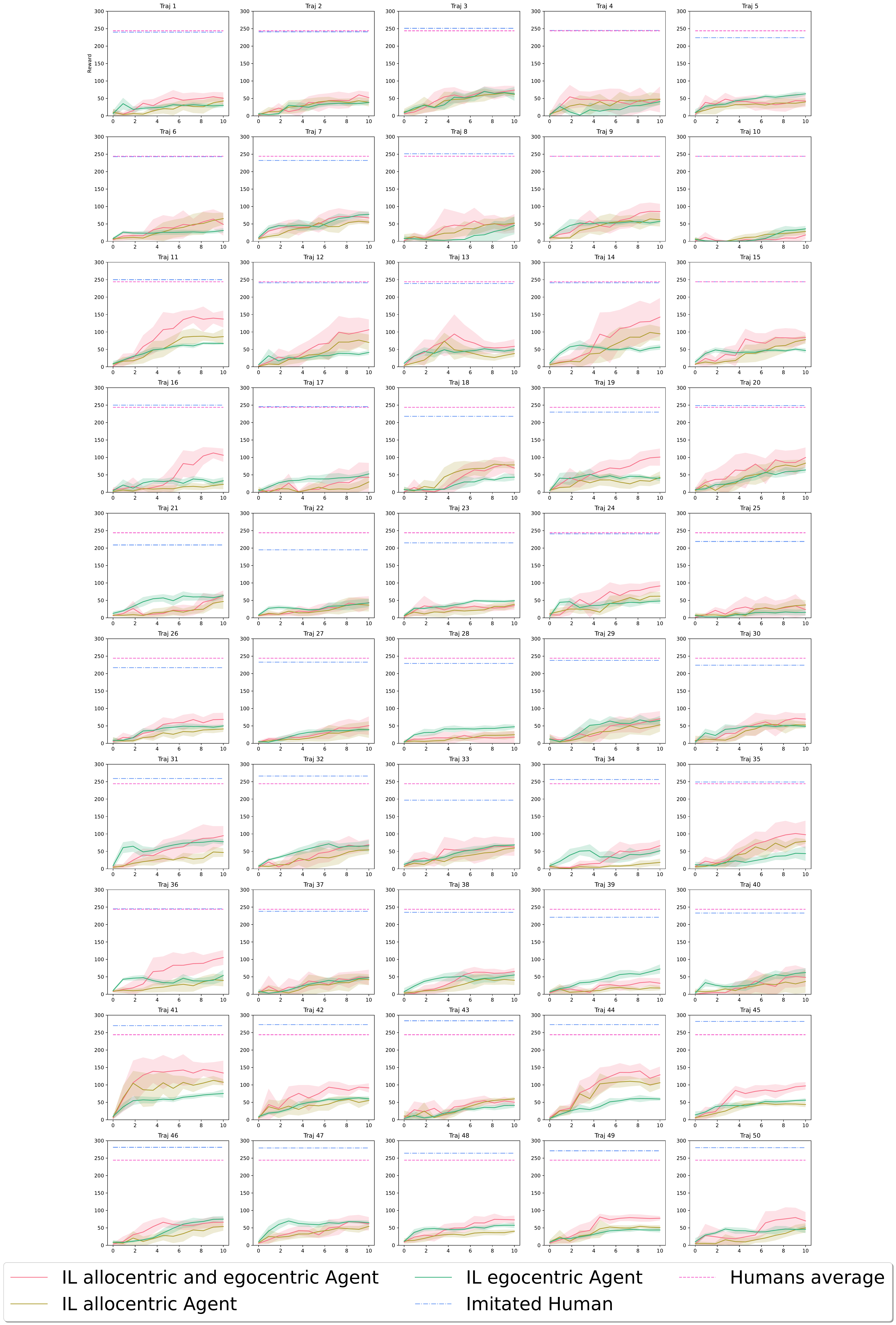}
    \caption{Comparison between IL with full state, allocentric only state and egocentric only state for all the $50$ human trajectories.}
    \label{fig:IL_ego}
\end{figure}

\begin{figure}
    \centering
    \includegraphics[width=0.8\linewidth,height = 20cm]{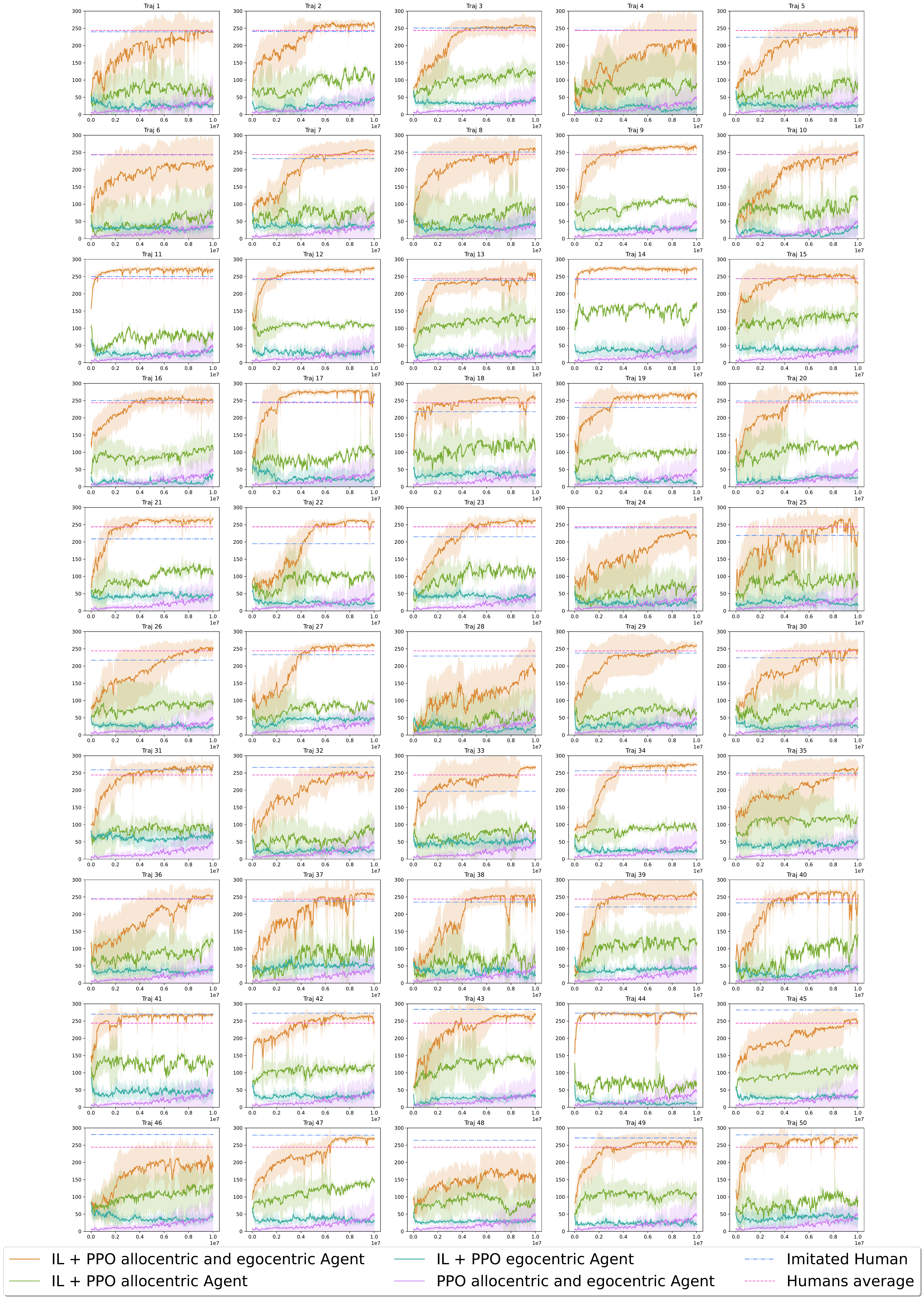}
    \caption{Comparison between IL+PPO with full state, allocentric only state and egocentric only state for all the $50$ human trajectories.}
    \label{fig:IL+RL_ego}
\end{figure}

\end{appendix}
\end{document}